%% file: main_v2.tex
\documentclass[11pt, letterpaper, logo, onecolumn, copyright, numbering]{minimax}
\usepackage{etoolbox}

\usepackage[authoryear, sort&compress, round]{natbib}

\usepackage[inkscapeformat=png]{svg}

\usepackage[most, breakable, skins]{tcolorbox}
\usepackage{academicons}

\tcbuselibrary{skins}
\usepackage{lipsum}
\usepackage{tabularx}
\usepackage{afterpage}
\usepackage{booktabs}
\usepackage{makecell}
\usepackage{multirow}
\usepackage{multicol} 
\usepackage{array}
\usepackage{float}
\usepackage{listings, listings-rust}
\usepackage{fontawesome5}
\usepackage{amssymb,graphicx}
\usepackage[dvipsnames]{xcolor}
\usepackage{hyperref}
\usepackage{cleveref}
\usepackage{longtable}
\usepackage{pdflscape}
\usepackage{adjustbox}
\usepackage{tikz}
\usetikzlibrary{calc,positioning,chains,shapes,arrows,fit,decorations.pathmorphing,patterns,fadings,shadows,patterns.meta,arrows.meta}
\usepackage{wrapfig}
\usepackage{dialogue}
\usepackage{algorithm}
\usepackage{algorithmic}
\usepackage{colortbl}
\usepackage{mdframed}
\usepackage{fontawesome5}
\usepackage{colortbl}
\usepackage{subcaption}
\usepackage{amssymb}
\usepackage{textcomp}
\usepackage{marvosym}

\usepackage{listings} 
\usepackage{xcolor}
\usepackage{CJKutf8}
\usepackage{tcolorbox} 
\usepackage[dvipsnames]{xcolor}
\usepackage{multicol}    

\usepackage{amsmath} 
\usepackage{cleveref}
\usepackage{hyperref}
\usepackage{graphicx}
\usepackage{multirow}

\input{math_commands.tex}

\input{showcase_format}

\theoremstyle{plain}

\theoremstyle{definition}

\theoremstyle{remark}

\usepackage{CJKutf8}

\definecolor{medgray55}{gray}{0.55}
\definecolor{medgray}{gray}{0.7}
\definecolor{litegray}{gray}{0.9}
\definecolor{gblue}{RGB}{210, 227, 252}
\definecolor{gred}{RGB}{250, 210, 207}
\definecolor{gyellow}{RGB}{254, 239, 195}
\definecolor{ggreen}{RGB}{206, 234, 214}
\definecolor{gorange}{RGB}{254, 223, 200}

\definecolor{gblue9}{RGB}{23, 78, 166}
\definecolor{gred9}{RGB}{165, 14, 14}
\definecolor{gyellow9}{RGB}{227, 116, 0}
\definecolor{ggreen9}{RGB}{13, 101, 45}
\definecolor{gorange9}{RGB}{176, 96, 0}

\definecolor{myblue}{rgb}{0,0,1}
\definecolor{myred}{rgb}{1,0,0}
\definecolor{mylightgray}{gray}{0.95}

\definecolor{highlightblue}{HTML}{185ABC}

\makeatletter

\renewcommand\paragraph{\@startsection{paragraph}{4}{\z@}%
            {-2.5ex\@plus -1ex \@minus -.25ex}%
            {1.25ex \@plus .25ex}%
            {\itshape\normalsize\bfseries}}
\makeatother
\newcolumntype{L}[1]{>{\raggedright\let\newline\\\arraybackslash\hspace{0pt}}m{#1}}
\newcolumntype{C}[1]{>{\centering}m{#1}}

\newcolumntype{R}[1]{>{\raggedleft\let\newline\\\arraybackslash\hspace{0pt}}m{#1}}

\definecolor{ao}{rgb}{0.0, 0.0, 1.0}

\newcommand\vcent[1]{\vcenter{\hbox{#1}}}

\newcommand\loudspeaker[1][3]{\ensuremath{\vcent{\rule{.6ex}{.6ex}}\kern-.5ex%
  \vcent{\scalebox{.6}[1]{\rotatebox[origin=center]{90}{$\blacktriangle$}}}%
  \ifnum#1>0\relax\kern.05ex\vcent{\scalebox{.4}{\ttfamily)}}%
  \ifnum#1>1\relax\kern-.4ex\vcent{\scalebox{.56}{\ttfamily)}}%
  \ifnum#1>2\relax\kern-.55ex\vcent{\scalebox{.7}{\ttfamily)}}%
  \fi\fi\fi}%
}

\definecolor{green}{rgb}{0.9,0.9,0.9}

\crefname{figure}{Fig.}{Figs.}
\crefname{appendix}{Appx.}{Appx.}
\crefname{table}{Tab.}{Tables}
\Crefname{table}{Tab.}{Tables}
\crefname{section}{Sec.}{Sec.}
\Crefname{section}{Sec.}{Sec.}
\crefname{equation}{Eq.}{Eqs.}
\Crefname{equation}{Eq.}{Eqs.}
\crefformat{equation}{Eq.~#2#1#3}
\Crefformat{equation}{Eq.~#2#1#3}
\crefname{paragraph}{Sec.}{Secs.}

\definecolor{blue_light_1}{RGB}{221, 232, 255}
\definecolor{blue_1}{RGB}{50, 126, 230}

\makeatletter
\renewcommand\subparagraph{%
 \@startsection {subparagraph}{5}{\z@ }{3.25ex \@plus 1ex
 \@minus .2ex}{-1em}{\normalfont \normalsize \bfseries }}%
\makeatother

\let\cite\citep

\makeatletter 
\renewcommand{\absfont}{\linespread{1.2}\fontsize{10}{12}\selectfont}
\makeatother  

\reportnumber{} 

\renewcommand{\today}{}

\author[*,1]{Full author list in Contributions\footnote{Please send correspondence to model@minimaxi.com.}}

\begin{abstract}
Reinforcement learning (RL) is becoming an important direction for post-training vision-language models (VLMs), but public training methodologies for unified multimodal RL remain much less mature, especially for heterogeneous reasoning and perception-heavy tasks.
We propose \textbf{V-Triune}, a \textbf{V}isual \textbf{Tri}ple \textbf{U}nified Rei\textbf{n}forcement L\textbf{e}arning methodology for unified multimodal RL.
It organizes training around three coordinated abstractions: \textit{Sample-Level Reward Routing}, \textit{Verifier-Level Outcome Verification}, and \textit{Source-Level Diagnostics}.
Within this methodology, Dynamic IoU provides localization-specific reward shaping that avoids reward ambiguity under loose thresholds and reward sparsity under strict ones.
Built on V-Triune, we develop Orsta (7B, 32B), a family of models jointly trained on eight reasoning and perception tasks.
Under matched budgets, unified training matches or outperforms specialist mixtures.
The final Orsta models improve over their backbones on MEGA-Bench, compare favorably with strong multi-task RL-VLM baselines, and transfer these gains to a broad set of downstream benchmarks.
These results show that unified RL can improve both reasoning and perception within a single VLM RL pipeline.
\end{abstract}

\begin{document}

\begin{figure}[t]
\centering
  \includegraphics[width=\textwidth]{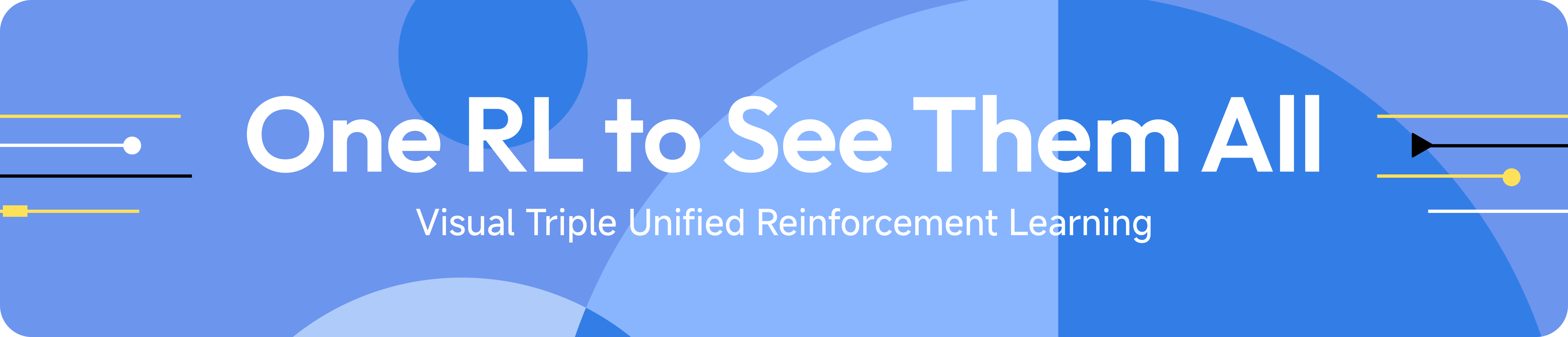}
  \label{fig:banner}
  \vspace{-10pt}
\end{figure}

\title{}
\maketitle

\input{sections_refined/1_introduction_v8.tex}
\input{sections_refined/6_related_work_v6.tex}
\input{sections_refined/3_method_v10.tex}
\input{sections_refined/5_experiment_v4.tex}

\input{sections_refined/7_conclusion_v1.tex}

\newpage
\section*{Contributions}

\noindent\textbf{Core Contributions}

Yan Ma$*$$^{1,4}$, Linge Du$*$$^{1,3}$, Xuyang Shen$*$$\dagger$$^{1}$, $^\textrm{\Letter}$Junjie Yan$^{1}$ 
\begingroup
\renewcommand\thefootnote{}\footnote{$*$ Equal Contribution; $\dagger$ Project Lead; $^\textrm{\Letter}$ Corresponding Author}
\addtocounter{footnote}{-1}
\endgroup

\noindent\textbf{Contributions}

Shaoxiang Chen$^{1}$, Pengfei Li$^{1}$, Qibing Ren$^{1,2}$

\noindent\textbf{Advisor}

Junjie Yan$^{1}$, Pengfei Liu$^{2,4}$, Yuchao Dai$^{3}$, Lizhuang Ma$^{2}$

\noindent\textbf{Affiliation}

$^{1}$ MiniMax

$^{2}$ Shanghai Jiao Tong University

$^{3}$ Northwestern Polytechnical University

$^{4}$ Generative Artificial Intelligence Lab (GAIR)

\newpage
\bibliography{new_reference}

\clearpage
\appendix
\input{sections/8_appx_model_and_data.tex}
\input{sections/4_infra.tex}

\input{sections/9_appx_metric_analysis.tex}
\input{sections/10_appx_common_downstream_tasks.tex}
\input{sections/11_appx_prompts.tex}
\input{sections/12_appx_reward_server.tex}
\input{sections/13_adaptive.tex}
\input{sections/14_appx_mega_bench_curves.tex}
\input{sections/15_appx_dynamic_iou_cases.tex}
\input{sections/16_appx_ovd_dynamic_iou.tex}

\end{document}

%% file: math_commands.tex
\usepackage{amsmath,amsfonts,bm}

\def\eqref#1{equation~\ref{#1}}

\def\1{\bm{1}}

\DeclareMathAlphabet{\mathsfit}{\encodingdefault}{\sfdefault}{m}{sl}
\SetMathAlphabet{\mathsfit}{bold}{\encodingdefault}{\sfdefault}{bx}{n}



%% file: showcase_format.tex
\definecolor{ababcol}{HTML}{F14738}
\definecolor{myhailuo2}{HTML}{F97669}
\definecolor{querycol}{HTML}{7964E8}
\definecolor{goldanswercol}{HTML}{FFB43B}
\definecolor{otherscol}{HTML}{FC5BCF}
\definecolor{myhailuo3light}{HTML}{FFA9FA}
\input{colors}
\tcbset{
    showcase/.style={
        fonttitle=\large,
        colback=white!20,  
        colframe=black,   
        coltitle=white,   
        boxrule=0.5mm,    
        arc=2mm,          
        outer arc=2mm,    
        left=1mm,         
        right=1mm,        
        top=1mm,          
        bottom=1mm,       
        width=\textwidth, 
        before skip=0.1pt,
        after skip=0.1pt,
    },
    context/.style={
        fontupper=\scriptsize,
        fonttitle=\large,
        colframe=querycol,     
        coltitle=white,   
        colback=white,    
        boxrule=0.3mm,    
        arc=2mm,          
        outer arc=2mm,    
        left=1mm,         
        right=1mm,        
        top=1mm,          
        bottom=1mm,       
        before skip=1pt,
        after skip=0.1pt, 
    },
    query/.style={
        fontupper=\scriptsize,
        fontlower=\scriptsize,
        colframe=querycol,     
        coltitle=white,   
        colback=white,    
        boxrule=0.1mm,    
        arc=2mm,          
        outer arc=2mm,    
        left=1mm,         
        right=1mm,        
        top=1mm,          
        bottom=1mm,       
        before skip=1pt,
        after skip=0.1pt,
    },
    abab/.style={
        fontupper=\scriptsize,
        fonttitle=,
        colframe=ababcol, 
        coltitle=white,   
        boxrule=0.5mm,    
        arc=2mm,          
        outer arc=2mm,    
        left=1mm,         
        right=1mm,        
        top=1mm,          
        bottom=1mm,       
        width=0.33\textwidth, 
        before skip=0.1pt,
        after skip=0.1pt, 
    },
    others/.style={
        fontupper=\scriptsize,
        colframe=myhailuo3light, 
        coltitle=white,
        boxrule=0.5mm,    
        arc=2mm,          
        outer arc=2mm,    
        left=1mm,         
        right=1mm,        
        top=1mm,          
        bottom=1mm,       
        width=0.33\textwidth, 
        before skip=0.1pt,
        after skip=0.1pt, 
    },
    goldanswer/.style={
        fontupper=\scriptsize,
        colframe=goldanswercol,     
        coltitle=white,   
        boxrule=0.5mm,    
        arc=2mm,          
        outer arc=2mm,    
        left=1mm,         
        right=1mm,        
        top=1mm,          
        bottom=1mm,       
        width=0.33\textwidth, 
        before skip=0.1pt,
        after skip=0.1pt, 
    },
}

%% file: colors.tex
\definecolor{myhailuo1dark}{HTML}{FC8900}
\definecolor{myhailuo2dark}{HTML}{F14738}
\definecolor{myhailuo3dark}{HTML}{D12AAA}
\definecolor{myhailuo4dark}{HTML}{4C4DC2}

\definecolor{myhailuo1}{HTML}{FFB43B}
\definecolor{myhailuo2}{HTML}{F97669}
\definecolor{myhailuo3}{HTML}{FC5BCF}
\definecolor{myhailuo4}{HTML}{7964E8}

\definecolor{myhailuo1light}{HTML}{FFD085}
\definecolor{myhailuo2light}{HTML}{FFA19F}
\definecolor{myhailuo3light}{HTML}{FFA9FA}
\definecolor{myhailuo4light}{HTML}{BDACFB}

\colorlet{myorange}{Orange!20}
\colorlet{mygreen}{LimeGreen!25}
\colorlet{myyellow}{Yellow!30}
\colorlet{myblue}{CornflowerBlue!25}
\colorlet{mybrown}{RawSienna!25}
\colorlet{mypurple}{Orchid!25}
\colorlet{myred}{Red!60}
\colorlet{myorangefull}{YellowOrange!60}
\colorlet{mybrownfull}{RawSienna!60}

\colorlet{myorangethick}{Orange!40}
\colorlet{mygreenthick}{LimeGreen!50}
\colorlet{myyellowthick}{Yellow!60}
\colorlet{mybluethick}{CornflowerBlue!50}

%% file: sections_refined/1_introduction_v8.tex
\section{Introduction}
\label{sec:intro}

Reinforcement learning (RL) is becoming an important direction for post-training foundation models~\citep{liu2025deepseek,xiao2026mimo}.
Compared with text-only settings, however, public and reproducible training methodologies for multimodal RL remain much less mature~\citep{qwen3.5,team2026kimi}.
Although more works have begun to explore RL post-training in visual settings~\citep{liu2025visual,ma2025deepperception,tan2025reason,liu2025othink,shen2025vlm,wang2025vl,yu2025perception,liu2025visionreasoner}, a stable methodology for scaling RL across heterogeneous visual tasks remains an important open problem for vision-language models (VLMs).

For VLMs, this challenge lies in handling both capability heterogeneity and verification heterogeneity within one pipeline.
Post-training must cover both reasoning-heavy and perception-heavy tasks, while their outcomes may require symbolic checking or continuous spatial feedback such as IoU.
A unified multimodal RL methodology must therefore support heterogeneous reward structures and verification regimes within the same optimization loop.
We study unified RL post-training for visual tasks with verifiable outcomes spanning both reasoning-heavy and perception-heavy settings.

Existing work usually addresses only part of this problem, focusing either on visual reasoning~\citep{huang2025vision,yang2025r1,meng2025mm,vl_rethinker} or on perception tasks such as detection and grounding~\citep{liu2025visual,ma2025deepperception,tan2025reason,liu2025othink,shen2025vlm,wang2025vl,yu2025perception,liu2025visionreasoner}.
A clear and stable RL framework for jointly improving a single VLM on both high-level reasoning and fine-grained perception remains missing.

In practice, unified multimodal RL breaks down for three concrete reasons.
First, different tasks, and even different samples, can require different reward compositions and verifiers, so hard-coding this logic into the trainer quickly becomes brittle and difficult to scale.
Second, for localization tasks such as detection and grounding, fixed IoU thresholds create two failure modes: loose thresholds blur reward differences between coarse predictions, whereas strict thresholds make rewards too sparse for stable learning.
Third, in joint training, aggregate metrics often hide source-specific instability, reward degradation, or other failures, making them difficult to detect and diagnose in time.

To address these blockers, we introduce \textbf{V}isual \textbf{Tri}ple \textbf{U}nified Rei\textbf{n}forcement L\textbf{e}arning (V-Triune), a training methodology for unified multimodal RL over heterogeneous visual tasks.
V-Triune organizes unified training around three coordinated abstractions: sample-level reward routing (\S\ref{subsec:sample_level_formatting}), which decouples reward composition and verifier choice from the trainer core; verifier-level outcome verification (\S\ref{subsec:verifier_level_reward_computation}), which provides a common interface for heterogeneous rewarding; and source-level diagnostics (\S\ref{subsec:metric_monitoring}), which expose source-specific failures hidden by aggregated metrics.
Within this methodology, Dynamic IoU Reward (\S\ref{subsec:dynamic_iou}) serves as a localization-specific reward-shaping mechanism by progressively increasing the IoU requirement, avoiding both reward ambiguity under loose thresholds and reward sparsity under strict ones.

Built on V-Triune, we develop the Orsta (\textbf{O}ne \textbf{R}L to \textbf{S}ee \textbf{T}hem \textbf{A}ll) model family at both 7B and 32B scales and jointly train these models on eight representative tasks under a single RL pipeline, including reasoning tasks such as mathematics, science, chart, and puzzle, as well as perception tasks such as detection, grounding, OCR, and counting.
Under matched budgets, unified training matches or outperforms specialist mixtures while remaining strong across both reasoning and perception within a single RL pipeline.
Orsta improves over its base models on MEGA-Bench~\citep{chen2024mega}, a comprehensive benchmark spanning over 440 diverse visual tasks, and these gains further transfer to downstream benchmarks.
Taken together, these results suggest that V-Triune provides an effective and practical training methodology for joint RL post-training of reasoning and perception in open VLMs.

Our main contributions are threefold.
First, we formulate unified VLM post-training as a heterogeneous multimodal RL problem and identify three practical blockers: rigid reward interfaces, the ambiguity-versus-sparsity trade-off in localization rewards, and lack of source-level observability during joint training.
Second, we propose V-Triune, a training methodology built on sample-level reward routing, verifier-level outcome verification, and source-level diagnostics, together with Dynamic IoU reward shaping for localization tasks.
Third, we instantiate this methodology in the Orsta family and validate it through eight-task unified training, matched-budget specialist comparisons, MEGA-Bench evaluation, broader reasoning and perception benchmarks, and extension to a new GUI task domain.

%% file: sections_refined/6_related_work_v6.tex
\section{Related Work}

RL post-training is becoming important for foundation models and is extending from language models to multimodal models~\citep{liu2025deepseek,qwen3.5,team2026kimi,team2025longcat,xiao2026mimo}.
However, public multimodal RL recipes remain much less mature than their text-only counterparts, especially for setups that span heterogeneous tasks and reward regimes.
As a result, most existing VLM-RL work is still organized around particular capability families or relatively homogeneous task settings, rather than a unified recipe for jointly training reasoning-heavy and perception-heavy tasks.

One major line of work focuses on multimodal reasoning.
Vision-R1~\citep{huang2025vision}, R1-OneVision~\citep{yang2025r1}, MM-Eureka~\citep{meng2025mm}, VL-Rethinker~\citep{vl_rethinker}, and GThinker~\citep{zhan2025gthinker} mainly target mathematics, science, and related reasoning tasks, typically using rule-based rewards and R1-style post-training to improve reasoning.
These works have substantially advanced multimodal reasoning RL, but their task composition and reward design remain centered on reasoning-heavy problems.
Another line focuses on perception-heavy or relatively homogeneous visual tasks.
Visual-RFT~\citep{liu2025visual}, DeepPerception~\citep{ma2025deepperception}, Reason-RFT~\citep{tan2025reason}, Perception-R1~\citep{yu2025perception}, VLM-R1~\citep{shen2025vlm}, and VisionReasoner~\citep{liu2025visionreasoner} show that RL can be effective for detection, grounding, OCR, counting, segmentation, and related tasks, often with task-specific rewards such as IoU or mAP.
Taken together, these two lines do not directly address how reasoning-heavy and perception-heavy tasks can stably coexist within the same RL training framework.

A smaller set of recent work moves toward broader multimodal RL settings.
Mixed-R1~\citep{xu2025mixedr1} studies multimodal RL with mixed answer types and reward forms, including open-ended text rewards, while OneThinker~\citep{feng2025onethinker} addresses heterogeneous image-video training through optimizer-side reward normalization and a GRPO~\citep{shao2024deepseekmath} variant.
These works move toward broader multimodal RL settings, but they address different problems from ours.
Our work instead studies a public methodology for jointly training open VLMs on reasoning-heavy and perception-heavy tasks, where capability heterogeneity, verifier heterogeneity, localization-specific reward shaping, and source-level observability must all be handled within one pipeline.


%% file: sections_refined/3_method_v10.tex
%
\begin{figure}[tb!]
    \centering
    \includegraphics[width=0.99\linewidth]{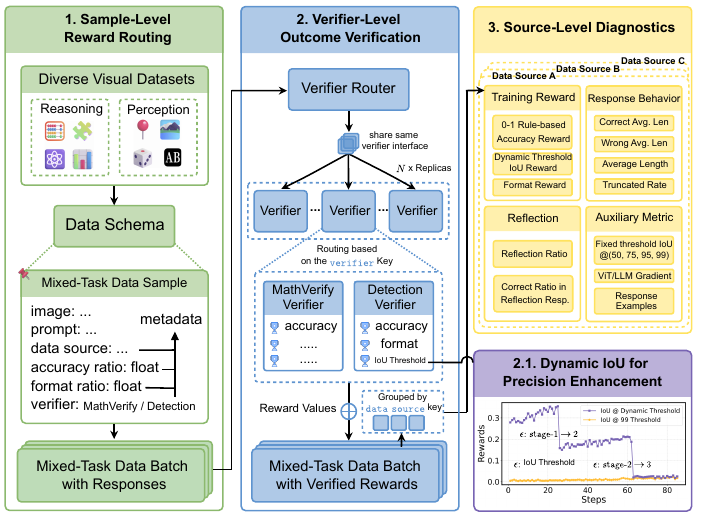}
    \caption{\textbf{Overview of V-Triune.} The methodology organizes unified multimodal RL around three training abstractions: sample-level reward routing, verifier-level outcome verification, and source-level diagnostics. They address rigid reward interfaces, localization reward ambiguity versus sparsity, and lack of source-level observability in mixed-task training.}
    \label{fig:framework}
    \vspace{-10px}
\end{figure}
\section{V-Triune: Visual Triple Unified Reinforcement Learning}
\label{sec:method}

This section presents V-Triune, our training methodology for unified multimodal RL in vision-language models.
The central problem is how to stably train reasoning-heavy and perception-heavy tasks within a single RL pipeline despite their different reward interfaces, outcome verification regimes, and observability requirements.

In practice, this setting breaks down in three recurring ways.
First, heterogeneous tasks do not fit a rigid shared reward interface, and hard-coded task-specific reward branches make the trainer brittle and difficult to extend.
Second, localization-centric tasks face an ambiguity-versus-sparsity trade-off under fixed IoU thresholds.
Third, aggregate-only monitoring can hide source-specific failure modes, making training problems difficult to localize.

V-Triune addresses these issues through three coordinated abstractions, shown in~\cref{fig:framework}.
Sample-level reward routing defines a clean boundary between the trainer and task-specific reward logic.
Verifier-level outcome verification provides a common interface for heterogeneous reward computation and incorporates Dynamic IoU for localization tasks.
Source-level diagnostics restore the observability that aggregate-only monitoring cannot provide.
Together, these abstractions make heterogeneous rewards, verification, and failure signals manageable within one pipeline, supporting stable unified training.
\subsection{Sample-Level Reward Routing}
\label{subsec:sample_level_formatting}

Unified multimodal RL needs an extensibility boundary between the trainer and task-specific reward logic.
Without such a boundary, adding new task families means inserting special-case reward branches into the training loop, which quickly becomes brittle and hard to maintain.
V-Triune makes this boundary explicit through sample-level reward routing.

Each sample carries a compact routing specification that determines which reward components should be used, how they are weighted, and which verifier should evaluate the rollout.
The trainer consumes only this shared routing interface, while task-specific verification logic stays outside the update loop.
This turns heterogeneous reward handling from trainer-side branching into a unified sample-side interface.

In practice, the routing metadata includes component weights, a verifier key, and the source identifier later used for diagnostics.
In this work, sample-level routing makes standard mixed-task RL explicit rather than introducing per-sample reward tuning.
The metadata only selects among a small set of verifier-defined reward templates, making explicit the routing that heterogeneous verifiable-reward training already requires.
The full schema is shown in~\cref{fig:data_format}.
What matters methodologically is that the trainer no longer needs task-specific reward code paths in order to mix reasoning and perception tasks in one run.

This design defines a clean extensibility boundary: adding a new task requires preparing data that matches an existing verification regime, or registering a new verifier, without rewriting the trainer itself.
The same routing metadata also supports source-aware diagnosis by allowing rewards and behaviors to be regrouped by source after verification.

\subsection{Verifier-Level Outcome Verification}
\label{subsec:verifier_level_reward_computation}

Sample-level routing determines which verification path a rollout should follow, but unified training still needs a common interface for heterogeneous outcome evaluation.
Some tasks can be checked with deterministic rules after parsing the model output, whereas others require continuous spatial scoring.
V-Triune handles this at the verifier level: the trainer sends predictions and references to the designated verifier, which performs parsing, verification, and reward computation through a shared interface.
This abstraction lets the trainer interact with heterogeneous reward logic in one way while preserving task-specific verification inside each verifier.
In this work we primarily instantiate two verifiers, corresponding to two verification regimes:
\noindent\paragraph{MathVerifyVerifier: Rule-verifiable outcomes}
This verifier handles tasks whose outputs can be parsed and deterministically checked, including mathematics, puzzles, science, chart reasoning, OCR, and counting.
For these tasks, we parse the model output into a normalized answer and verify it against the reference using \texttt{math\_verify}~\citep{mathverify2025}.
The resulting accuracy reward follows the standard 0--1 rule-based form:
\begin{equation}
    R_\text{acc}(\hat{a}, a) = \mathbb{I}(\texttt{verify}(\texttt{parse}(\hat{a}), \texttt{parse}(a)))
    \label{eq:math_rule}
\end{equation}
where $\hat{a}$ denotes the predicted answer and $a$ the ground-truth answer.
In our setup, model responses are instructed to place the final answer inside \texttt{\textbackslash boxed\{\}}.

\noindent\paragraph{DetectionVerifier: Localization-centric outcomes}
This verifier handles tasks such as detection and grounding, where outputs must satisfy both a structural format and a spatial accuracy criterion.
We therefore compute a composite reward with separate format and accuracy terms.
To enforce the required output structure, we define a format reward as
\begin{equation}
    R_{\text{format}}(o_q) = 0.25 \sum_{i=1}^{4} \mathbb{I}(\text{count}(o_q, s_i) = 1)
    \label{eq:format_rule}
\end{equation}
where $o_q$ is the model response to query $q$, and the four format tags are $\{s_i\}_{i=1}^4 =$\{\textit{<think>}, \textit{</think>}, \textit{<answer>}, \textit{</answer>}\}.

For the spatial component, we use IoU-based accuracy:
\begin{equation}
R_\text{acc}(\hat{a}, a) =
\begin{cases}
\text{IoU}(\hat{a}, a), & \text{if}\quad\text{IoU}(\hat{a}, a) \ge \epsilon \\
0, & \text{else}
\end{cases},
\qquad \text{where} \quad
\text{IoU}(\hat{a}, a) = \frac{\text{Area}(\hat{a} \cap a)}{\text{Area}(\hat{a} \cup a)}
\label{eq:iou_rule}
\end{equation}
where $\hat{a}$ is the predicted box and $a$ is the ground-truth box.
The final reward combines the two parts as $\alpha_{\text{acc}} \cdot R_{\text{acc}} + \alpha_{\text{format}} \cdot R_{\text{format}}$, where $\alpha_{\text{acc}}$ and $\alpha_{\text{format}}$ are specified by the sample-level routing metadata.

Verifier-level computation gives different task families a clear boundary for reward logic while keeping the trainer unchanged.
For new task domains that can reuse an existing output structure and verification regime, only the data routing needs to be updated.
We verify this point with the GUI-domain extension experiment in~\cref{subsec:gui_extension}.

\subsubsection{Dynamic IoU for Precision Enhancement}
\label{subsec:dynamic_iou}

Within \textit{DetectionVerifier}, the main optimization challenge is to construct a localization reward that is informative without becoming too sparse.
IoU is the most direct spatial signal for detection and grounding, but a fixed threshold creates a practical dilemma.

With a loose threshold such as IoU@50~\citep{yu2025perception}, reward density is high but reward ambiguity remains severe: multiple coarse boxes can receive nearly identical rewards, leaving limited pressure for finer localization.
With a highly strict threshold such as near-exact matching, the objective becomes clearer but early rollouts receive almost no positive feedback, creating a cold-start problem.
This issue is amplified in unified training because localization rewards must coexist with sharper rule-based rewards from reasoning tasks.

To address this trade-off, we introduce Dynamic IoU, a dense-to-strict reward shaping mechanism.
Rather than fixing one threshold, we use a simple three-stage schedule that begins with denser supervision, then progressively raises the IoU requirement, and finally enforces near-exact localization later in training.
This staging preserves learnability in early optimization while still imposing high-precision supervision in the late phase.

Dynamic IoU makes high-precision localization learnable under joint training by combining dense early supervision with stricter late-stage refinement.
We provide the exact thresholds and stage boundaries in the implementation details, and compare this design against fixed thresholds and alternative schedules in the experiments and appendix.
\subsection{Source-Level Diagnostics}
\label{subsec:metric_monitoring}

Even with sample-level routing and verifier-level computation, unified RL can still fail in ways that aggregate metrics do not expose.
Global averages may remain seemingly stable while individual sources undergo reward collapse, format drift, abnormal token generation, local task suppression, or optimization instability.
V-Triune therefore tracks verified samples at the source level using the same \texttt{data\_source} key introduced in the routing interface.

We organize these signals into four groups: training reward, response behavior, reflection-related metrics, and auxiliary diagnostics.
Training reward metrics provide a source-level breakdown of rule-based accuracy, format reward, and IoU-based reward.
Response behavior metrics include average length, lengths of correct and incorrect responses, and truncation rates, which help reveal verbosity drift or collapse.
Reflection metrics provide a lightweight online proxy for response strategy by tracking whether reflective cues appear in responses; the precise definition is given in~\cref{appx:reflection_metrics}.
Auxiliary diagnostics include fixed-threshold IoU, ViT/LLM gradients, and response examples for qualitative analysis.

In our experiments, these diagnostics directly informed two training decisions in the final recipe: freezing the vision encoder after observing ViT gradient explosion, and filtering leaked image special tokens before reward recomputation~(\cref{sec:training_recipe}).
They also reveal divergent response-length and reflection patterns across task families and help detect when one source is being overshadowed by stronger training signals elsewhere.
Source-level diagnostics are therefore part of the training methodology rather than a dashboard add-on: they make unified training actionable and diagnosable.

%% file: sections_refined/5_experiment_v4.tex
\section{Experiment}
\label{sec:experiment}

\begin{table*}[t]
\centering
\caption{Fair-budget evidence under a fixed training budget. Left: MEGA-Bench task-composition curves for the 7B off-policy setting. Right: benchmark breakdown at the same 60-step budget. Unified = Reason+Perception. MME-R = MME-Reasoning~\citep{yuan2025mme}. CharXiv(RQ) = CharXiv Reasoning Question. COCO (S$\mid$M) reports single-object and multi-object mAP in COCO val-2017. COCO mAP uses the standard cocoapi; see details in~\cref{sec:coco_evaluation}. Best results are bolded and second-best results are underlined.}
\label{fig:fair_budget_evidence}
\begin{minipage}[t]{0.31\textwidth}
\vspace{-1pt}
\centering
\includegraphics[width=\linewidth]{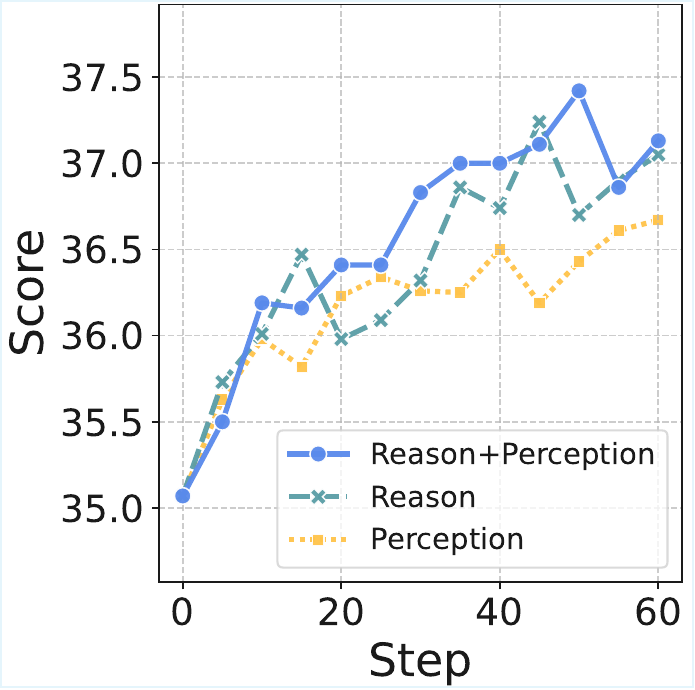}
\end{minipage}\hfill
\begin{minipage}[t]{0.68\textwidth}
\vspace{0pt}
\centering
\scriptsize
\setlength{\tabcolsep}{4pt}
\resizebox{\linewidth}{!}{%
\begin{tabular}{lccccc}
\toprule
\multicolumn{6}{c}{Reasoning Benchmarks} \\
\midrule
Mixture & MMMU & MathVista & MathVision & MME-R & CharXiv (RQ) \\
\midrule
Unified & \underline{56.56} & \underline{71.40} & \underline{30.51} & \textbf{28.20} & \textbf{44.50} \\
Reason-only & 54.89 & \textbf{71.70} & \textbf{31.24} & \underline{28.03} & \underline{44.00} \\
Perception-only & \textbf{56.67} & 68.20 & 29.59 & 26.26 & 41.00 \\
\bottomrule
\end{tabular}
}

\vspace{4pt}

\resizebox{\linewidth}{!}{%
\begin{tabular}{lccccc}
\toprule
\multicolumn{6}{c}{Perception Benchmarks} \\
\midrule
Mixture & HrBench4K & VStar & COCO (S$\mid$M) & OCRBenchV2 & ScreenSpot-Pro \\
\midrule
Unified & \underline{73.75} & \textbf{82.20} & \underline{80.69} $\mid$ \textbf{38.62} & \textbf{55.87} & \underline{23.91} \\
Reason-only & 73.25 & 78.53 & 78.36 $\mid$ 33.87 & \underline{55.77} & \textbf{23.97} \\
Perception-only & \textbf{76.00} & \underline{80.10} & \textbf{80.78} $\mid$ \underline{37.37} & 55.33 & 23.78 \\
\bottomrule
\end{tabular}
}
\end{minipage}
\end{table*}
\vspace{-5pt}

\subsection{Experimental Setup}
\label{subsec:implementation details}

We evaluate V-Triune at two levels.
First, under a fixed budget, we compare a unified mixture against two specialist mixtures: \textit{Reason-only} and \textit{Perception-only}.
Second, we report the final Orsta models under a longer training horizon to evaluate overall competitiveness on broad and task-specific benchmarks.

We use Qwen2.5-VL-7B and 32B~\citep{Qwen2.5-VL} as backbones.
We study unified RL post-training on strong public VLM backbones when the target outcomes are verifiable.
RL post-training uses a 47.7K-sample VQA training set with verifiable answers from 18 sources, covering four reasoning tasks (math, puzzle, science, and chart) and four perception tasks (detection, grounding, counting, and OCR).
Detailed data construction is given in~\cref{subsec:data,tab:data_source}.
Appendix~\cref{sec:curated_vs_random} further reports a matched-count curated-versus-random control to separate data quality from the training methodology.
We use GRPO~\citep{shao2024deepseekmath} in all main experiments, with a rollout batch size of 1024 and 8 sampled responses per prompt, and implement the unified RL pipeline on top of Verl~\citep{sheng2024hybridflow}.
Controlled comparisons use a fixed training budget, whereas final-model results use 3 training epochs.
For stability, we freeze the vision encoder and update only the LLM layers.
Rollouts are generated with vLLM using \texttt{temperature=1.0}, \texttt{top-p=1.0}, and \texttt{max\_length=2048}, and localization tasks use a three-stage Dynamic IoU schedule.

\subsection{Does Unified Training Help Under Fair Budgets?}
\label{subsec:fair_budget}

We first study unified training under a fixed training budget.
Unless otherwise stated, all fair-budget comparisons use the same 7B off-policy with 8 optimization steps per rollout.
We compare the unified mixture against two specialist mixtures under the same budget.

The left panel of~\cref{fig:fair_budget_evidence} provides the first evidence through the MEGA-Bench task-composition curves.
Under the same 60-step budget, \textit{Reason+Perception} follows the strongest or tied-strongest trajectory throughout training.
This suggests that joint training does not introduce a clear reasoning-versus-perception trade-off on MEGA-Bench.

We further compare the same three mixtures on a 10-benchmark suite spanning both reasoning and perception, and report the fixed-budget benchmark breakdown in~\cref{fig:fair_budget_evidence}.
The benchmark list and evaluation protocols are given in~\cref{sec:eval_details}.
The unified mixture remains stronger or comparable on both sides and reaches the best result on multiple benchmarks.
Across this suite, the unified model matches or exceeds the specialist baselines on 5/10 benchmarks and stays close on the rest.

This is consistent with improved budget efficiency under unified training, since the unified model remains competitive despite receiving less task-specific exposure per task.
One plausible explanation is mild cross-task regularization, where exposure to both reasoning and
perception tasks improves shared learning signals beyond specialist training alone.
We leave a deeper mechanistic analysis of this effect to future work.


\begin{table*}[t]
\centering

\begin{minipage}[t]{0.29\linewidth}
\vspace{0pt}
\centering

\captionof{table}{Performance on MEGA-Bench.}
\label{tab:mega_bench_core}

\small
\setlength{\tabcolsep}{2pt}

\begin{tabular}{@{}lr@{}}
\toprule
Model & Score \\
\midrule
\multicolumn{2}{c}{7B Models} \\
\midrule
Qwen2.5-VL-7B      & 35.07 \\
MM-Eureka-7B       & 35.96 \\
VL-Rethinker-7B    & 37.25 \\
\textbf{Orsta-7B}  & \textbf{38.31} \\
$\Delta$ (Ours - Backbone) & +3.24 \\
\bottomrule
\end{tabular}

\vspace{5pt}

\begin{tabular}{@{}lr@{}}
\toprule
\multicolumn{2}{c}{32B Models (0321)} \\
\midrule
Qwen2.5-VL-32B-0321     & 11.87 \\
MM-Eureka-32B           & 18.57 \\
VL-Rethinker-32B        & 19.41 \\
\textbf{Orsta-32B-0321} & \textbf{25.94} \\
$\Delta$ (Ours - Backbone) & +14.07 \\
\bottomrule
\end{tabular}

\vspace{5pt}

\begin{tabular}{@{}lr@{}}
\toprule
\multicolumn{2}{c}{32B Models (0326)} \\
\midrule
Qwen2.5-VL-32B-0326     & 43.67 \\
Gemma3-27B              & 41.82 \\
InternVL-3-38B          & \textbf{46.69} \\
\textbf{Orsta-32B-0326} & 45.77 \\
$\Delta$ (Ours - Backbone) & +2.10 \\
\bottomrule
\end{tabular}

\end{minipage}
\hfill
\begin{minipage}[t]{0.68\linewidth}
\vspace{0pt}
\centering

\captionof{table}{Comparison against strong multi-task RL-VLM baselines on reasoning and perception benchmarks.}
\label{tab:baseline_comparison}

\scriptsize
\setlength{\tabcolsep}{3pt}

\resizebox{\linewidth}{!}{%
\begin{tabular}{@{}lccccc@{}}
\toprule
\multicolumn{6}{c}{\textbf{Reasoning Benchmarks}} \\
\midrule
Model & MMMU & MathVista & MathVision & MME-R & CharXiv (RQ) \\
\midrule
\textbf{Orsta-7B}       & \textbf{57.10} & 72.50 & 31.73 & \textbf{31.14} & \textbf{48.40} \\
MM-Eureka-7B            & 55.33          & 74.10 & 30.84 & 28.45          & 42.10 \\
VL-Rethinker-7B         & 56.70          & \textbf{75.40} & \textbf{32.46} & 29.38 & 44.00 \\
VisionReasoner-7B       & 56.56          & 69.70 & 29.20 & 25.84          & 41.20 \\
\bottomrule
\end{tabular}
}

\vspace{5pt}

\resizebox{\linewidth}{!}{%
\begin{tabular}{@{}lccccc@{}}
\toprule
\multicolumn{6}{c}{\textbf{Perception Benchmarks}} \\
\midrule
Model & HrBench4K & VStar & COCO (S$\mid$M) & OCRBenchV2 & ScreenSpot-Pro \\
\midrule
\textbf{Orsta-7B}       & \textbf{77.25} & \textbf{81.68} & \textbf{80.73} $\mid$ \textbf{41.41} & \textbf{56.05} & 23.91 \\
MM-Eureka-7B            & 59.62          & 57.07          & 79.73 $\mid$ 35.84                   & 53.38          & 24.23 \\
VL-Rethinker-7B         & 65.12          & 68.60          & 72.50 $\mid$ 31.54                   & 55.70          & \textbf{24.48} \\
VisionReasoner-7B       & 74.38          & 80.63          & 80.22 $\mid$ 36.58                   & 55.44          & 24.23 \\
\bottomrule
\end{tabular}
}

\vspace{10pt}

\captionof{table}{Extension to a new GUI task domain with 3K ShowUI samples under the same 60-step fixed-compute budget.}
\label{tab:gui_extension}
\vspace{4pt}

\small
\setlength{\tabcolsep}{20pt}

\resizebox{0.99\linewidth}{!}{%
\begin{tabular}{@{}lcc@{}}
\toprule
Mixture & ScreenSpot-Pro & OCRBenchV2 \\
\midrule
Unified           & 23.91         & 55.87 \\
Perception-only   & 23.78         & 55.33 \\
Perception-only + GUI-3K          & 29.85         & 55.96 \\
Unified + GUI-3K             & \textbf{31.68} & \textbf{56.09} \\
\bottomrule
\end{tabular}
}

\end{minipage}

\end{table*}

\subsection{General Performance on MEGA-Bench}
\label{subsec:mega_bench}

The fixed-budget results in~\cref{subsec:fair_budget} show that unified RL training matches or outperforms specialist mixtures under matched budgets.
We now turn to the final Orsta models and report their overall performance on MEGA-Bench, which spans over 440 diverse tasks for general VLM capability.
For each backbone, we train both on-policy and off-policy variants for 3 epochs, designate one model as Orsta based on its MEGA-Bench performance, and evaluate that same selected model on all subsequent benchmarks.
We provide the full MEGA-Bench evaluation curves in~\cref{appx:mega_bench_curves}.

Overall, Orsta consistently improves the MEGA-Bench performance of its base model at both 7B and 32B scales.
For the 7B model, unified RL post-training yields a clear overall gain.
The same pattern holds at 32B, but with two different gain profiles in~\cref{tab:mega_bench_core}: for the \texttt{0321} checkpoint,\footnote{\texttt{0321} and \texttt{0326} follow the release dates of the public HuggingFace checkpoints. The former shows noticeably weaker perception and formatting abilities, while the latter is a stronger later release.} RL post-training brings a much larger improvement; for the stronger \texttt{0326} checkpoint, V-Triune still delivers stable and notable gains.

\begin{figure*}[t]
\centering
\includegraphics[width=\linewidth]{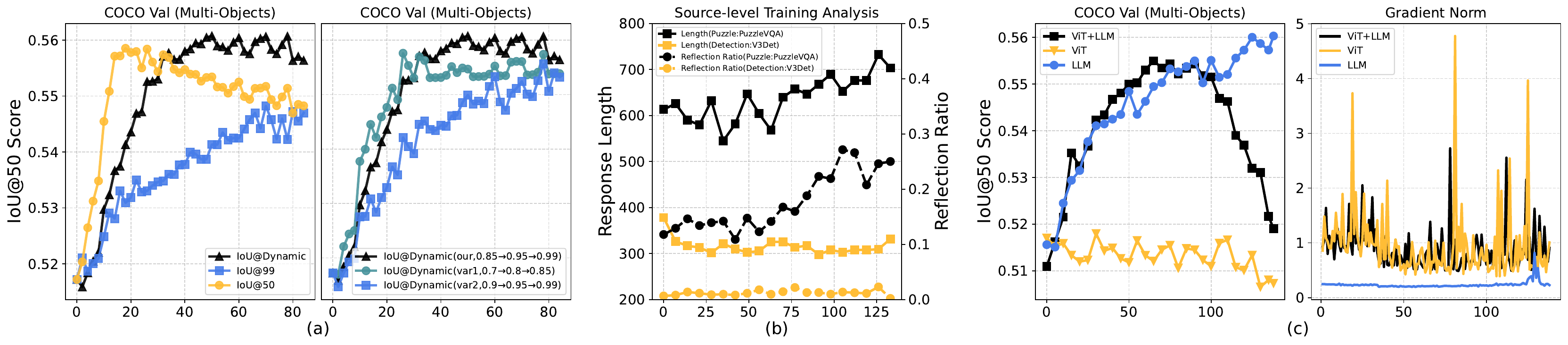}
\caption{Dynamic IoU ablations and source-level diagnostics. (a) Dynamic IoU ablations on COCO multi-object using only the detection and grounding subsets for training. The left plot compares fixed IoU@50, fixed IoU@99, and the main Dynamic IoU schedule; the right plot compares alternative staged schedules. Fixed IoU@50 learns quickly but degrades later, whereas fixed IoU@99 is more stable but learns more slowly; the main Dynamic IoU schedule gives the best balance. (b) Source-level logs from full 8-task Orsta-32B unified RL training show that the reasoning-heavy Puzzle source has increasing response length and reflection ratio, whereas the localization-centric Detection source remains short and nearly reflection-free. (c) In a full 8-task 7B training ablation that updates the ViT only, the LLM only, or both, continuing to update the vision encoder hurts COCO performance and increases gradient norms, motivating the decision to freeze the vision encoder and connector in the main experiments. More detailed failure analyses are provided in~\cref{sec:training_recipe}.}
\label{fig:combined_iou_source_level}
\end{figure*}

\subsection{Comparison with Strong Multi-task RL-VLM Baselines}
\label{subsec:baseline_comparison}

Beyond the overall MEGA-Bench gains in~\cref{subsec:mega_bench}, we compare Orsta-7B against three strong multi-task RL-VLM baselines on the same 10-benchmark suite as in~\cref{subsec:fair_budget}.
MM-Eureka-7B~\citep{meng2025mm} and VL-Rethinker-7B~\citep{vl_rethinker} are more reasoning-oriented, whereas VisionReasoner-7B~\citep{liu2025visionreasoner} is more perception-centric.

As shown in~\cref{tab:baseline_comparison}, Orsta-7B achieves the best score on 7/10 benchmarks.
On the reasoning side, it wins MMMU, MME-Reasoning, and CharXiv(RQ), while staying competitive on MathVista and MathVision.
On the perception side, it reaches the top score on 4/5 benchmarks, with the clearest margin on the challenging COCO multi-object setting.
This complements the fair-budget evidence in~\cref{subsec:fair_budget}: the benefit of unified training remains visible in the final model and is not confined to either reasoning or perception alone.

\subsection{Why is Dynamic IoU Necessary for Localization Tasks?}
\label{subsec:dynamic_iou_ablation}

We next examine why Dynamic IoU is necessary for localization-centric tasks and why a staged schedule is effective.
Unless otherwise stated, the ablations in this subsection use only the detection and grounding subsets of the training data in the same 7B off-policy setup.
The main schedule uses IoU thresholds $0.85$, $0.95$, and $0.99$ over the first 10\%, the next 15\%, and the remaining training steps.

On COCO multi-object, the fixed-threshold comparison in~\cref{fig:combined_iou_source_level}(a) shows the two failure modes directly.
IoU@50 learns quickly but degrades later, which we attribute to reward ambiguity under a loose threshold.
Qualitative cases in~\cref{appx:dynamic_iou_cases} support this interpretation: late-stage predictions often drift among multiple coarse boxes that remain similarly rewarded under IoU@50, rather than continuing to sharpen around the ground-truth box.
IoU@99 is more precise but learns much more slowly because the reward is too sparse early on.
Dynamic IoU avoids both issues by remaining learnable early while enforcing higher precision later.
We further compare alternative staged schedules in~\cref{fig:combined_iou_source_level}(a).
The loose variant ($0.7 \rightarrow 0.8 \rightarrow 0.85$) plateaus because its final threshold remains too permissive, whereas the strict variant ($0.9 \rightarrow 0.95 \rightarrow 0.99$) slows early learning by making the initial reward too sparse.
Our main schedule ($0.85 \rightarrow 0.95 \rightarrow 0.99$) gives the best balance.

The same advantage pattern also appears on the OVDEval negation subset; full curves are given in~\cref{appx:ovd_dynamic_iou}.
We also implemented an adaptive scheduler based on the batch-level bbox success rate, but it does not yield a clear advantage over the fixed three-stage schedule in our setting; full results are given in~\cref{appx:adaptive}.

\subsection{What Do Source-Level Diagnostics Reveal?}
\label{subsec:source_level_diagnostics}

These diagnostics reveal signals aggregate metrics can hide: global averages may look stable while source-specific response behavior and failure modes diverge during training~(\cref{fig:combined_iou_source_level}).

On the behavior side, \cref{fig:combined_iou_source_level}(b) uses Puzzle and Detection as two representative sources and shows unified training does not collapse them into the same response pattern.
For the reasoning-heavy Puzzle source, both response length and reflection ratio increase over training.
For the localization-centric Detection source, responses remain short and direct, and reflection stays low.
This suggests unified training preserves task-appropriate response patterns across task families.
Combined with the fair-budget results, these diagnostics suggest that joint training improves robustness across task families without forcing them into a uniform response pattern.
More detailed multi-source dynamics are provided in~\cref{sec:training_dynamics_analysis}.

On the stability side, \cref{fig:combined_iou_source_level}(c) shows that continuing to update the vision encoder quickly hurts COCO performance and increases total gradient norms.
\cref{sec:training_recipe} further shows that this instability appears as gradient explosion in the ViT and soon hurts detection and grounding performance.
The same source-level logs also exposed leaked image special tokens in model responses, which can cause the number of visual features to no longer match the input sequence and directly lead to training failure.
Based on these findings, we freeze the vision encoder and connector in the main experiments and filter leaked image special tokens before reward recomputation; more detailed layer-wise gradients, image-token failure cases, and stability analysis are provided in~\cref{sec:training_recipe}.

\subsection{Can V-Triune Extend to New Task Domains?}
\label{subsec:gui_extension}

Finally, we test whether V-Triune can absorb a new task domain beyond the original training mix, and whether the benefit of unified training remains after the new domain is added.
As a concrete case, we add about 3K GUI grounding samples from ShowUI~\citep{lin2025showui}, a domain not covered in the original dataset.
We treat GUI grounding as a new domain to test the scalability of our framework, which can be added by reusing the existing localization-style verifier regime.

As shown in~\cref{tab:gui_extension}, adding ShowUI data substantially improves ScreenSpot-Pro, with the unified model rising from 23.91 to 31.68.
Notably, after adding the same GUI data, \textit{Unified + GUI} still outperforms \textit{Perception + GUI} by +1.83 points on ScreenSpot-Pro.
Reasoning-heavy data therefore does not block learning in the new GUI domain, and unified training continues to show positive transfer once the model has basic in-domain visual support.

These results suggest that V-Triune can extend beyond the original training domains while preserving the benefit of unified training.
In the ShowUI case, this extension is straightforward because the GUI task can reuse the existing localization-style verifier regime.

%% file: sections_refined/7_conclusion_v1.tex
\section{Discussion \& Future Work}

We presented V-Triune, a training methodology for unified multimodal RL over reasoning-heavy and perception-heavy VLM tasks.
By organizing training around reward routing, verifier-level outcome verification, and source-level diagnostics, together with Dynamic IoU for localization-centric tasks, V-Triune addresses rigid reward interfaces, localization reward ambiguity versus sparsity, and lack of observability in mixed-task RL.
Under matched budgets, unified training matches or outperforms specialist mixtures, and the final Orsta models improve over their backbones on MEGA-Bench and a broad set of downstream benchmarks, while the same recipe also extends to a new GUI domain.

Two future directions seem especially important.
One is to extend unified RL from static benchmark settings to multimodal agentic tool-use tasks.
The other is to generalize this training recipe beyond vision, toward joint RL training across speech, video, and text.

%% file: sections/8_appx_model_and_data.tex
\clearpage
\section{Data Curation}
\label{subsec:data}
\begin{table}[h]
\caption{Data source composition and curation. The curated corpus contains 27,133 reasoning samples (Math, Puzzle, Science, Chart; 56.8\%) and 20,633 perception samples (Detection, Grounding, Counting, OCR; 43.2\%), for a total of 47,766 examples (47.7K).}
\centering
\small
{  
{
\begin{tabular}{@{}L{1.2cm} L{2cm} L{3.0cm} r r L{2.1cm}@{}}
\toprule
Task & Count (Proportion) & Data source name &
\multicolumn{1}{c}{After curation} &
\multicolumn{1}{c}{Original count} &
Notes \\
\midrule
\multirow{3}{*}{Math} & \multirow{3}{*}{11,810 (24.72\%)} &
mm math & 3,539 & 5,901 & \\
& & geometry3k & 2,539 & 3,002 & \\
& & mmk12 & 5,732 & 15,616 & \\
\midrule
\multirow{2}{*}{Puzzle} & \multirow{2}{*}{5,980 (12.52\%)} &
PuzzleVQA + AlgoPuzzleVQA & $2{,}648 \times 2$ & $3{,}800 \times 2$ &
\tiny{Puzzle data are duplicated because the original puzzle data size is relatively small.} \\
& & VisualPuzzles & $342 \times 2$ & $1{,}168 \times 2$ &
\tiny{Puzzle data are duplicated because the original puzzle data size is relatively small.} \\
\midrule
\multirow{3}{*}{Science} & \multirow{3}{*}{4,339 (9.08\%)} &
ScienceQA & 536 & 4,114 & \\
& & SciVQA & 1,264 & 15,120 & \\
& & ViRL39K (``STEM'' \& ``Science'') &
2,539 & 4,431 & \\
\midrule
\multirow{4}{*}{Chart} & \multirow{4}{*}{5,004 (10.48\%)} &
ChartQAPro & 498 & 1,948 & \\
& & ChartX & 2,353 & 4,848 & \\
& & Table-VQA-Bench & 496 & 1,500 & \\
& & ViRL39K (Tables / Diagrams / Charts) & 1,657 & 6,189 & \\
\midrule
\multirow{2}{*}{Detection} & \multirow{2}{*}{8,000 (16.75\%)} &
V3Det & 4,000 & 15,000 &
\tiny{We randomly sample a 15k subset from 183,354 images; after filtering we obtain
6,287 samples and then randomly select 4k.} \\
& & Object365 & 4,000 & 15,000 &
\tiny{We randomly sample a 15k subset from 1.74M images; after filtering we obtain
8,889 samples and then randomly select 4k.} \\
\midrule
Grounding & 4,870 (10.20\%) & D$^3$ & 4,870 & 20,278 & \\
\midrule
Count & 1,725 (3.61\%) & CLEVR & 1,725 & 4,000 &
\tiny{We sample a 4k subset from the full 35k dataset.} \\
\midrule
\multirow{2}{*}{OCR} & \multirow{2}{*}{6,038 (12.64\%)} &
LLaVA-OneVision (OCR-en) & 3,092 & 8,000 &
\tiny{We sample 8k images from the 56,613 images in the \texttt{ocr\_vqa}
category of LLaVA-OneVision-Mid-Data.} \\
& & EST-VQA & 2,946 & 8,000 &
\tiny{We sample 7k images from the 17,047 images in the EST-VQA training set.} \\
\bottomrule
\end{tabular}}
} 
\label{tab:data_source}
\end{table}

We select four reasoning tasks—Math, Puzzle, Science, and Chart—for their varied reasoning demands, and four perception tasks—Detection, Grounding, Counting, and OCR—for their broad coverage of visual understanding. Data sources for each task are listed below:

\begin{itemize}
    \item For the Math task, \texttt{mm\_math}~\citep{sun2024mm}, \texttt{geometry3k}~\citep{lu2021inter}, and \texttt{mmk12}~\citep{meng2025mm} are chosen.%
    \item For the Puzzle task, \texttt{PuzzleVQA}\citep{chia2024puzzlevqa} and \texttt{AlgoPuzzleVQA}\citep{ghosal2024algopuzzlevqa} are merged due to their shared origin, and \texttt{VisualPuzzles}~\citep{song2025visualpuzzles} is additionally included.%
\item For the Science task, \texttt{ScienceQA}~\citep{lu2022learn}, \texttt{SciVQA}~\citep{borisova2025scivqa}, and the ``Broader STEM Topics'' and ``(GradeSchool) Science'' categories from \texttt{ViRL39K}~\citep{wang2025vl} are used.%
    \item For the Chart task, \texttt{ChartQAPro}~\citep{masry2025chartqapro}, \texttt{ChartX}~\citep{xia2024chartx}, \texttt{Table-VQA}~\citep{kim2024tablevqa}, and the Tables/Diagrams/Charts categories from \texttt{ViRL39K}~\citep{wang2025vl} are used.%
    \item For the Detection task, \texttt{V3Det}~\citep{xie2023described} and \texttt{Object365}~\citep{shao2019objects365} are chosen.%
    \item For the Grounding task, \texttt{$D^3$}~\citep{xie2023described} is used.%
    \item For the Counting task, \texttt{CLEVR}~\citep{johnson2017clevr,tan2025reason} is used.%
    \item For the OCR task, English OCR questions are extracted from \texttt{LLaVA-OV Data}~\citep{li2024llava} and \texttt{EST-VQA}~\citep{wang2020general}.%
\end{itemize}

\begin{figure}[htb!]
    \centering
    \includegraphics[width=0.98\linewidth]{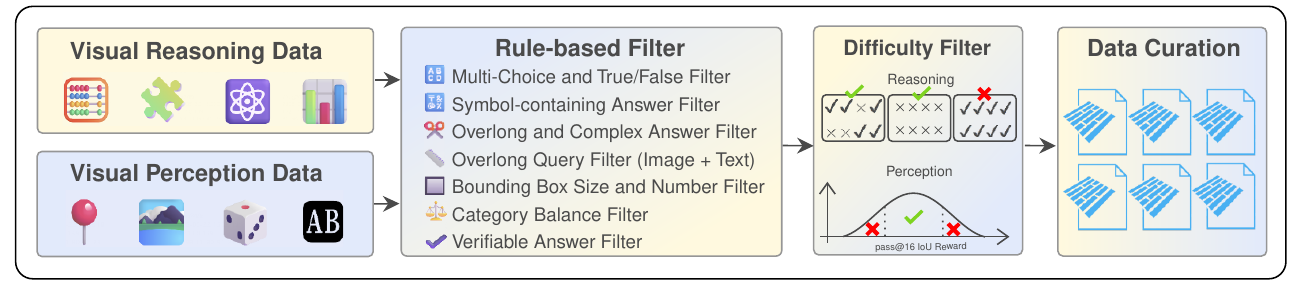}
    \caption{\textbf{Data Curation Process}. First, visual reasoning and visual perception data pass through a rule-based filter, which removes samples that do not meet preset criteria. Subsequently, the data enters a difficulty filter, which removes samples that are too easy or too hard based on model performance, ultimately producing the Curated Dataset.}
    \label{fig:data_curation}
\end{figure}

To reduce noise, we apply a two-stage data filtering process (Figure~\cref{fig:data_curation}): (1) rule-based filtering and (2) difficulty-based filtering. This yields 47.7K high-quality samples across 18 datasets and 8 tasks. To mitigate dataset bias, puzzle data is duplicated to ensure sufficient coverage. The final corpus includes approximately \textbf{20.6K perception} and \textbf{27.1K reasoning} samples, primarily consisting of single-image, single-turn conversations.

\paragraph{First Stage: Rule-based Filter} 
For four visual reasoning tasks, the following filters are applied:
\begin{itemize}
    \item Multiple-choice and true/false questions that are prone to hacking are discarded.~\citep{kimiteam2025kimivltechnicalreport}
    \item Answers containing symbols such as ``='', ``['', ``]'', ``('', ``)'', and ``;'' are removed, as the absence of these symbols may cause answer mismatches even if the numeric values are correct.
    \item Answers longer than 20 characters are discarded to avoid overly complex answers.
\end{itemize}

The filtering process for visual perception tasks involves additional complexity:
\begin{itemize}
    \item \textbf{Detection:} Following \texttt{Qwen2.5-VL}~\citep{Qwen2.5-VL}, data is converted to relative coordinates. Single-box samples contain one box per category, while multi-box samples retain original annotations. Samples with over 10 boxes per category or boxes exceeding 50\% of the image are removed. A 1:2 single-to-multi-box ratio is enforced, and category-level long tails are avoided.
    \item \textbf{Grounding:} Data is processed into relative coordinates, and data with a box size greater than 50\% of the image is discarded. Complex phrase labels are filtered out.%
    \item \textbf{Counting:} Data is balanced per category and only English data is retained.%
    \item \textbf{OCR:} Only English OCR data is retained, and final labels must be verifiable by \texttt{math\_verify}~\citep{mathverify2025}. Since no verifiable reward model (RM) is designed, the OCR task data must pass this validation.
\end{itemize}

\paragraph{Second Stage: Difficulty-based Filter}

To remove low-value samples, easy questions already solvable by the base model are filtered out.

For reasoning tasks, we use \texttt{Qwen2.5-VL-32B-0321} to compute pass@8, retaining only samples with $0 < \text{pass@8} < 100\%$.
For perception tasks, specifically detection and grounding, pass@16 is computed using \texttt{Qwen2.5-VL-7B} with a 0.5 IoU threshold, and samples with cumulative IoU rewards between 2 and 10 are selected.
This split keeps reasoning filtering based on a stronger reasoning-oriented model, while perception filtering follows the reward setting used for localization-style supervision.

All curated data is stored in Parquet format~\citep{parquet} and uniformly mixed for training without online filtering or curriculum scheduling. 

%% file: sections/4_infra.tex
\clearpage
\section{Insights from Source-Level Monitoring}
\label{sec:training_recipe}

Source-level monitoring was essential for turning unified RL into a stable training pipeline.
It exposed several concrete issues that were difficult to see from aggregate metrics alone, especially vision-encoder instability and leaked image special tokens.
This section summarizes the corresponding adjustments used in our final recipe.

\subsection{Stabilizing Training by Freezing the Vision Encoder}

\begin{figure}[htb!]
    \centering
    \includegraphics[width=0.98\linewidth]{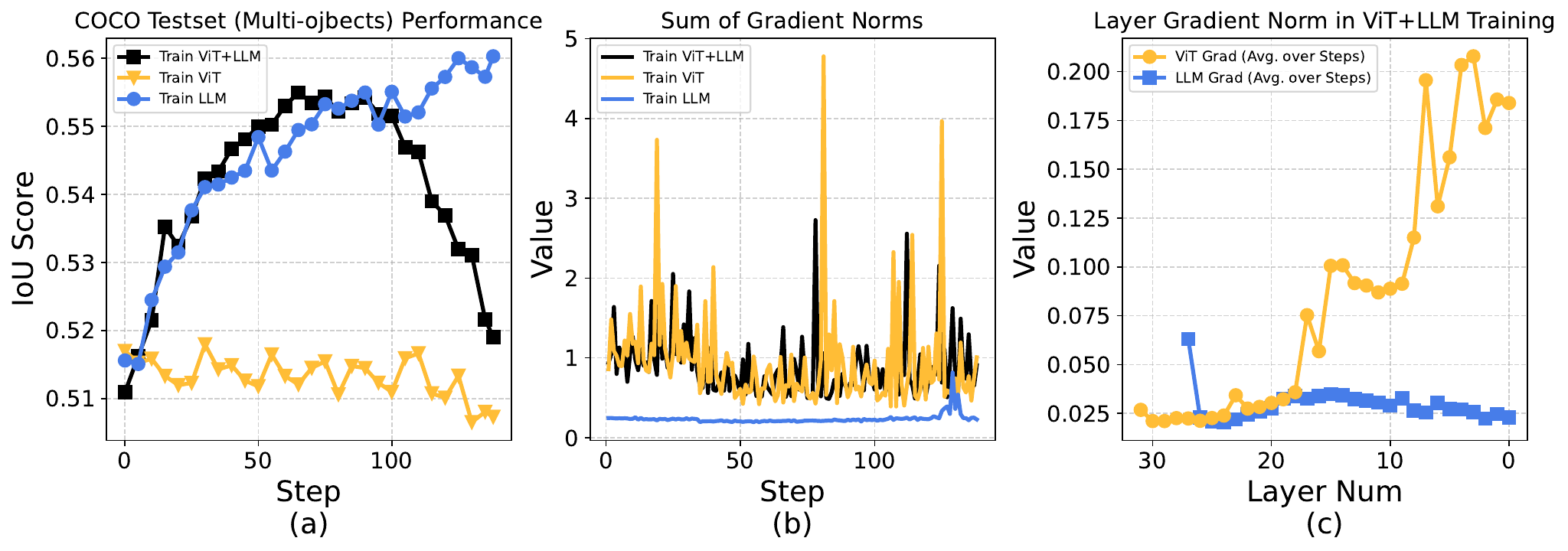}
    \caption{\textbf{Analysis of vision-encoder training instability.} (a) COCO performance under different training schemes. (b) Sum of gradient norms under different training schemes. (c) Layer-wise gradient norms of the vision encoder and LLM during full-parameter training. Updating the vision encoder leads to a performance drop and much less stable gradients, while the LLM remains comparatively stable.}
    \label{fig:disable_vit}
\end{figure}

In initial experiments, we performed full-parameter training by jointly optimizing the vision encoder and LLM.
However, detection performance consistently collapsed after several dozen steps, regardless of hyperparameter settings.
Log analysis revealed unusually large and spiking gradient norms, suggesting instability originating from the vision encoder.
To verify this, we compared three training configurations: (1) LLM-only, (2) vision encoder-only, and (3) full-parameter training, all under identical RL settings on Orsta-7B with mixed-task data.
We monitored COCO performance, total gradient norm, and layer-wise gradient trends during full-parameter training.

As shown in~\cref{fig:disable_vit}a, joint training leads to a performance drop, whereas LLM-only training maintains stable gains.
Vision encoder-only training yields minimal improvement, indicating that the main RL gains do not come from updating the vision encoder.
\Cref{fig:disable_vit}b shows that training the vision encoder produces much larger gradient norms than LLM-only training.

Layer-wise analysis in~\cref{fig:disable_vit}c confirms this pattern: LLM gradients remain relatively stable across layers, while vision-encoder gradients amplify during backpropagation.
This gradient explosion destabilizes training and undermines visual performance.
We therefore freeze the vision encoder and connector in the main experiments.

The precise cause of this instability remains open.
For the present work, the practical conclusion is that continuing to update the vision encoder is not beneficial under our unified RL setup, whereas freezing the vision encoder and connector gives substantially more stable training.

\subsection{Mitigating Leaked Image Special Tokens}

\begin{figure*}[ht]
  \centering
  \begin{tcolorbox}[title=Sample,label={box:sample}]
  \includegraphics[width=\linewidth]{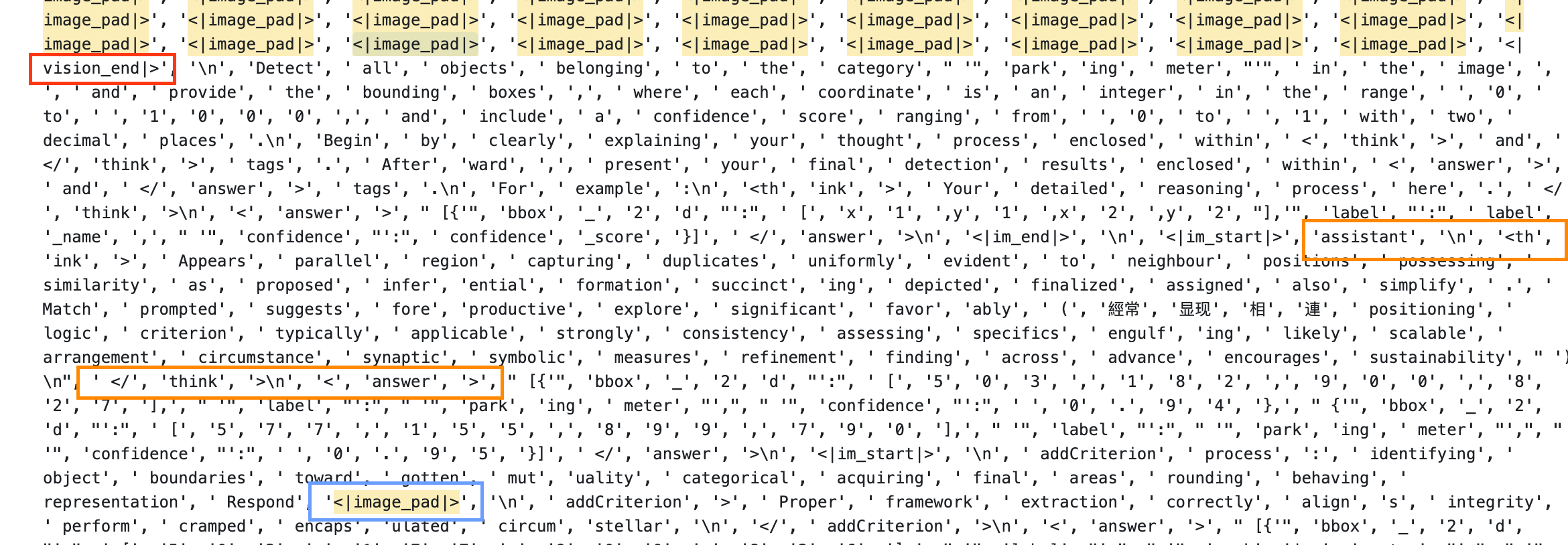}
  \end{tcolorbox}
  \caption{\textbf{An Example of Leaked Image Tokens in VLM Outputs.}}
  \label{fig:image_token}
\end{figure*}

To enable accurate advantage estimation, logits for both the query and the generated response are recomputed, as those returned by the inference engine may be imprecise.
During the forward pass, image placeholders (highlighted in the red box in~\cref{fig:image_token}, appearing before the ``vision\_end'' token)
are replaced with visual features extracted by the vision encoder and connector.
However, the model may mistakenly generate special tokens (highlighted in the blue box in~\cref{fig:image_token}), such as image or video placeholders, that lack corresponding features.
These leaked tokens cause the number of visual features to no longer match the input sequence and can directly lead to training failure.
We therefore filter all such special tokens from the rollout sequence before reward recomputation.

%% file: sections/9_appx_metric_analysis.tex
\clearpage
\section{Training Dynamics Analysis}
\label{sec:training_dynamics_analysis}
\begin{figure}[h!]
    \centering
    \includegraphics[width=0.90\linewidth]{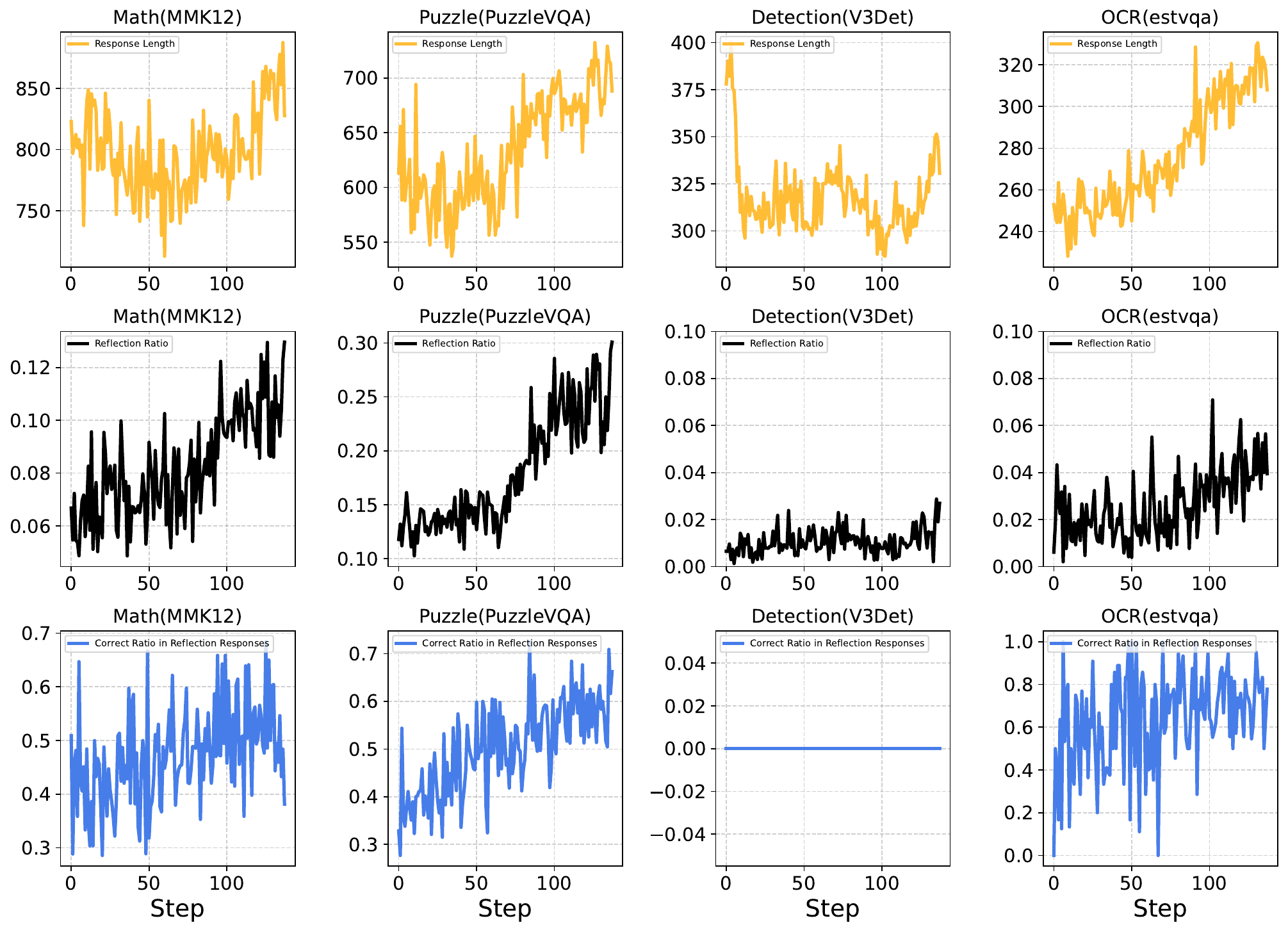}
    \caption{Training dynamics of response length (top row), reflection ratio (middle row), and correct ratio in reflection responses (bottom row) during training steps for Math (MMK12), Puzzle (PuzzleVQA), Detection (V3Det), and OCR (estvqa) tasks using the Orsta-32B-0321 off-policy setting. Each column corresponds to a different task, and each row represents a distinct metric.}
    \label{fig:train_analysis}
\end{figure}

This appendix provides supplementary source-level dynamics for four representative tasks: Math (MMK12), Puzzle (PuzzleVQA), Detection (V3Det), and OCR (estvqa), all drawn from Orsta-32B-0321 off-policy training logs. We report three metrics defined in~\cref{subsec:metric_monitoring}: response length, reflection ratio, and correctness of reflection responses. A more detailed explanation of the reflection-related metrics is provided in~\cref{appx:reflection_metrics}.

As shown in~\cref{fig:train_analysis}, response behavior varies substantially across tasks. Reasoning-oriented tasks such as Math and Puzzle exhibit increasing response length and reflection usage over training, whereas Detection remains shorter and shows near-zero reflection. OCR displays a different trajectory from Detection, underscoring that perception tasks are not behaviorally uniform.

The bottom row further shows that reflection quality also differs across tasks. Math, Puzzle and OCR exhibit improving correctness in reflection responses, and Detection stays near zero throughout. We include these plots as supplementary evidence for the source-level behavior divergence discussed in the main text.

%% file: sections/10_appx_common_downstream_tasks.tex
\clearpage
\section{Benchmark Details}
\label{sec:eval_details}

\label{subsec:bench_details}
We evaluate general VLM capability with \textbf{MEGA-Bench}~\citep{chen2024mega} and report its scores using the official evaluation code.
For the 10-benchmark suite used in the fixed-budget comparison and the strong-baseline comparison, we evaluate \textbf{MMMU}~\citep{yue2024mmmu}, \textbf{MathVista}~\citep{lu2023mathvista}, \textbf{MathVision}~\citep{mathvision}, \textbf{MME-Reasoning}~\citep{yuan2025mme}, \textbf{CharXiv} (RQ)~\citep{wang2024charxiv}, \textbf{HrBench4K}~\citep{hrbench}, \textbf{VStar}~\citep{vstar}, \textbf{COCO}~\citep{lin2014microsoft}, \textbf{OCRBenchV2}~\citep{fu2024ocrbench}, and \textbf{ScreenSpot-Pro}~\citep{li2025screenspot}.
Among them, \textbf{OCRBenchV2} is evaluated with Lmms-eval~\citep{zhang2024lmmsevalrealitycheckevaluation}, and the other eight benchmarks except \textbf{COCO} are evaluated with VLMEvalKit~\citep{duan2024vlmevalkit}.
For \textbf{COCO}, we use the official \texttt{cocoapi}; the detailed evaluation procedure is given in~\cref{sec:coco_evaluation}.
All bounding boxes and keypoints are represented using coordinate values relative to the original input image dimensions.

\clearpage
\section{Evaluation on COCO}
\label{sec:coco_evaluation}

We conduct our evaluation on the COCO val-2017 dataset~\citep{lin2014microsoft}, which contains 4,952 images with 36,781 ground-truth bounding boxes. The dataset includes 593 images with a single object (593 boxes) and 4,359 images with multiple objects (36,188 boxes). For the experiment, we use the instruction shown in~\cref{fig:det_sample} to prompt the model to generate a list of all target detections for each of the 4,952 images. The model operates at a temperature of 0 and outputs all bboxes in the format: [\{`bbox\_2d': [x1,y1,x2,y2],`label': label\_name\} ...] at one time.

The model's output boxes are parsed into the COCO format, and we use the official \texttt{cocoapi} to calculate the mean Average Precision (mAP). The mAP computation requires a confidence score for each prediction to rank them. We use the predicted box's area relative to the total image area as a pseudo-confidence score. The score is calculated as follows:
$$
\text{score} = \frac{(x_2 - x_1) \times (y_2 - y_1)}{\text{image\_width} \times \text{image\_height}}
$$

To validate the robustness of our evaluation, we also conducted an ablation study on the choice of the pseudo-confidence function. We implemented and compared several alternative heuristics, including methods based on object position (\texttt{center\_bias}) and shape (\texttt{aspect\_ratio}), alongside \texttt{fixed} and \texttt{random} baselines. As shown in~\cref{tab:confidence_ablation}, the mAP scores are remarkably stable across all deterministic heuristics, with a total spread of around 0.5 mAP. This stability suggests that our evaluation results are not sensitive to the specific choice of the ranking method. Therefore, we adopt the simple and interpretable \texttt{area\_ratio} method for all main experiments reported in this paper.

\begin{table}[hb!]
\centering
\caption{Ablation study on pseudo-confidence scoring methods for Qwen2.5-VL-7B-Instruct on the full COCO val-2017 dataset.}
\label{tab:confidence_ablation}
\begin{tabular}{lc}
\toprule
\textbf{Scoring Method} & \textbf{mAP@50:95} \\
\midrule
\texttt{area\_ratio} (our choice) & 33.63 \\
\texttt{center\_bias}             & 33.60 \\
\texttt{aspect\_ratio}            & 33.08 \\
\texttt{fixed} (1.0)              & 33.07 \\
\midrule
\texttt{random} (baseline)        & 33.05 \\
\bottomrule
\end{tabular}
\end{table}

\clearpage
\section{Additional Controls}
\label{sec:additional_controls}

\subsection{Matched-Count Curated-vs.-Random Control}
\label{sec:curated_vs_random}

To isolate the effect of data quality from the rest of the training pipeline, we compare the main Orsta-7B model trained on the curated 47.7K corpus against a matched-count random control. The random control uses the exact same backbone, RL recipe, and training budget as Orsta-7B; the only change is that each data source is replaced by a stratified random subset from the corresponding raw pool, while preserving the same post-curation sample count. This keeps the task and source distribution fixed and varies only the data quality.

\begin{table*}[ht]
\centering
\small
\setlength{\tabcolsep}{3.5pt}
\caption{Matched-count curated-vs.-random control on the 10-benchmark suite. Curated-47.7K consistently improves over the random control on complex reasoning and fine-grained perception benchmarks.}
\label{tab:curated_vs_random}
\resizebox{\linewidth}{!}{
\begin{tabular}{lcccccccccc}
\toprule
Model & MMMU & MathVista & MathVision & MME-R & Charxiv (RQ) & HrBench4K & VStar & COCO (M) & OCRBenchV2 & ScreenSpot Pro \\
\midrule
Orsta-7B-Random   & 56.33 & 72.20 & 31.57 & 28.62 & 46.40 & 74.38 & \textbf{83.25} & 40.38 & 55.83 & 23.85 \\
Orsta-7B-Curated  & \textbf{57.10} & \textbf{72.50} & \textbf{31.73} & \textbf{31.14} & \textbf{48.40} & \textbf{77.25} & 81.68 & \textbf{41.41} & \textbf{56.05} & \textbf{23.91} \\
Gain              & +0.77 & +0.30 & +0.16 & +2.52 & +2.00 & +2.87 & -1.57 & +1.03 & +0.22 & +0.06 \\
\bottomrule
\end{tabular}
}
\end{table*}

As shown in~\cref{tab:curated_vs_random}, the random control is already competitive, which indicates that the unified V-Triune pipeline remains effective even without curation. At the same time, the curated corpus yields additional gains on more demanding reasoning and fine-grained perception benchmarks, suggesting that reward-aware filtering improves data efficiency beyond simply increasing sample count.

\subsection{Detection Fast Path and Latency Profiling}
\label{sec:fast_path_latency}

We also profile the end-to-end inference cost of the detection prompt used in the main experiments and compare it against a shorter direct-mode prompt that removes the explicit CoT trigger. Both measurements are conducted on COCO val-2017 with Orsta-7B, using a single H200 GPU and vLLM v0.11.0 under greedy decoding.

\begin{table}[ht]
\centering
\small
\caption{Direct-mode fast path for detection on COCO val-2017. Removing the explicit CoT trigger shortens responses, improves throughput, and slightly improves mAP.}
\label{tab:fast_path_latency}
\begin{tabular}{lccc}
\toprule
Mode & Avg. Tokens & FPS & mAP@50:95 \\
\midrule
Standard CoT Prompt & 208.0 & 25 & 33.63 \\
Direct Prompt       & 93.5  & 30 & 34.40 \\
\bottomrule
\end{tabular}
\end{table}

The direct-mode result complements the source-level behavior analysis in the main text. For detection, the model does not require a long reasoning-style response at inference time: removing the explicit CoT trigger reduces output length, improves throughput, and slightly improves COCO performance.

%% file: sections/11_appx_prompts.tex
\clearpage
\section{Query Example of Detection and Grounding}
\begin{figure}[htb!]
  \centering
  \begin{tcolorbox}[title=Query Example of Detection and Grounding]
  \begin{lstlisting}
Please detect all instances of the following category within the image:
{LABEL}. 

Let's think step by step and output the final answer in <answer> and </answer> tags.
For example:
Your detailed reasoning process here.
<answer>
[{'bbox_2d': [x1,y1,x2,y2],'label': label_name}]
</answer>
  \end{lstlisting}
  \end{tcolorbox}
  \caption{\textbf{Example query format for detection and grounding tasks.} The query instructs VLMs to identify instances of a given object and format the output in a specific reasoning-answer format.}
  \label{fig:det_sample}
\end{figure}
%

%% file: sections/12_appx_reward_server.tex
%
%
%
\newpage
\section{Sample-level Data Scheme for Unified Training}
\label{appx:data_schema}
\begin{figure}[tb!]
    \centering

    \begin{tcolorbox}[title=Data Format, label={box:data_format}]
    \begin{lstlisting}
{
    "data_source": Value(dtype="string"),
    "images": Sequence(feature=Image(mode=None, decode=True)),
    "prompt": [
        {
            "content": Value(dtype="string"),
            "role": Value(dtype="string")
        }
    ],
    "ability": Value(dtype="string"),
    "reward_model": {
        "answer": Value(dtype="string"),
        "ground_truth": Value(dtype="string"),
        "accuracy_ratio": Value(dtype="float32"),
        "format_ratio": Value(dtype="float32"),
        "verifier": Value(dtype="string"),
        "verifier_parm": Value(dtype="dict")
    },
    "extra_info": {
        "id": Value(dtype="string"),
        "image_path": Value(dtype="string")
    }
}
    \end{lstlisting}
    \end{tcolorbox}
    \caption{\textbf{Sample-level Data Scheme for Unified Training}. This format, implemented using HuggingFace datasets, allows fine-grained control over reward computation by defining \texttt{reward\_model} (including reward types, weights like \texttt{accuracy/format\_ratio}) and \texttt{verifier} specifications at the individual sample level. This enables flexible and scalable handling of diverse multimodal tasks.}
    \label{fig:data_format}
\end{figure}

\section{Detailed Explanation of Reflection Metrics}
\label{appx:reflection_metrics}

This appendix provides a detailed breakdown of the reflection metrics used in our source-level metric monitoring. These metrics are designed to quantitatively assess the model's self-correction and reasoning processes.

\paragraph{Reflective Word Set}

Following \citet{ma2025rethinking}, we track a curated list of 15 English words and phrases that indicate a reflective or self-correcting thought process. A response is considered "reflective" if it contains one or more of the following terms:
\begin{itemize}
    \item re-check, re-evaluate, re-examine, re-think
    \item recheck, reevaluate, reexamine, rethink
    \item reevaluation
    \item check again, think again, try again
    \item verify, wait, yet
\end{itemize}

\paragraph{Metric Definitions}
Based on this word set, we define two metrics:

\begin{enumerate}
    \item \textbf{Reflection Ratio ($R_{\text{reflect}}$)}: This metric measures the overall frequency of reflective responses. It is defined as the total number of responses containing at least one reflective word ($N_{\text{reflect}}$) divided by the total number of all responses ($N_{\text{total}}$).
    \begin{equation}
        R_{\text{reflect}} = \frac{N_{\text{reflect}}}{N_{\text{total}}}
    \end{equation}

    \item \textbf{Correctness Rate within Reflection ($C_{\text{reflect}}$)}: This metric assesses the effectiveness of the model's reflective reasoning. It is defined as the number of reflective responses that are also correct ($N_{\text{correct\_reflect}}$) divided by the total number of reflective responses ($N_{\text{reflect}}$).
    \begin{equation}
        C_{\text{reflect}} = \frac{N_{\text{correct\_reflect}}}{N_{\text{reflect}}}
    \end{equation}
\end{enumerate}

We emphasize that this keyword-based approach serves as a rough proxy for reflective behavior, not a precise measurement. It is intended for lightweight, online monitoring to gauge general trends in the model's reasoning process, rather than for a formal or rigorous evaluation of its reflection capabilities.

%% file: sections/13_adaptive.tex
\clearpage
\section{Adaptive IoU Threshold Scheduling}
\label{appx:adaptive}

To compare our fixed three-stage Dynamic IoU schedule against an adaptive alternative, we implement a controller that automatically adjusts the IoU threshold according to the model's current batch-level success rate.
All runs in this section are trained on the detection and grounding subsets of our data, and we monitor IoU@50 on the OVDEval negation subset throughout training.

We define a discrete set of thresholds $\mathcal{S} = \{0.5, 0.55, 0.6, \dots, 0.95, 0.99\}$ and initialize training at $T_0 = 0.5$.
At each training step $t$, we compute the \textbf{Batch Success Rate (BSR)}, defined as the proportion of predicted boxes in the current batch whose IoU exceeds the current threshold $T_t$.
The threshold for the next step, $T_{t+1}$, is updated according to a target-success hyperparameter $\tau$:

\begin{equation}
T_{t+1} =
\begin{cases}
\text{next}(T_t, \mathcal{S}) & \text{if } \text{BSR} > \tau \\
\text{prev}(T_t, \mathcal{S}) & \text{if } \text{BSR} < \tau \\
T_t & \text{otherwise}
\end{cases}
\end{equation}

\begin{figure}[h!]
    \centering
    \includegraphics[width=0.80\linewidth]{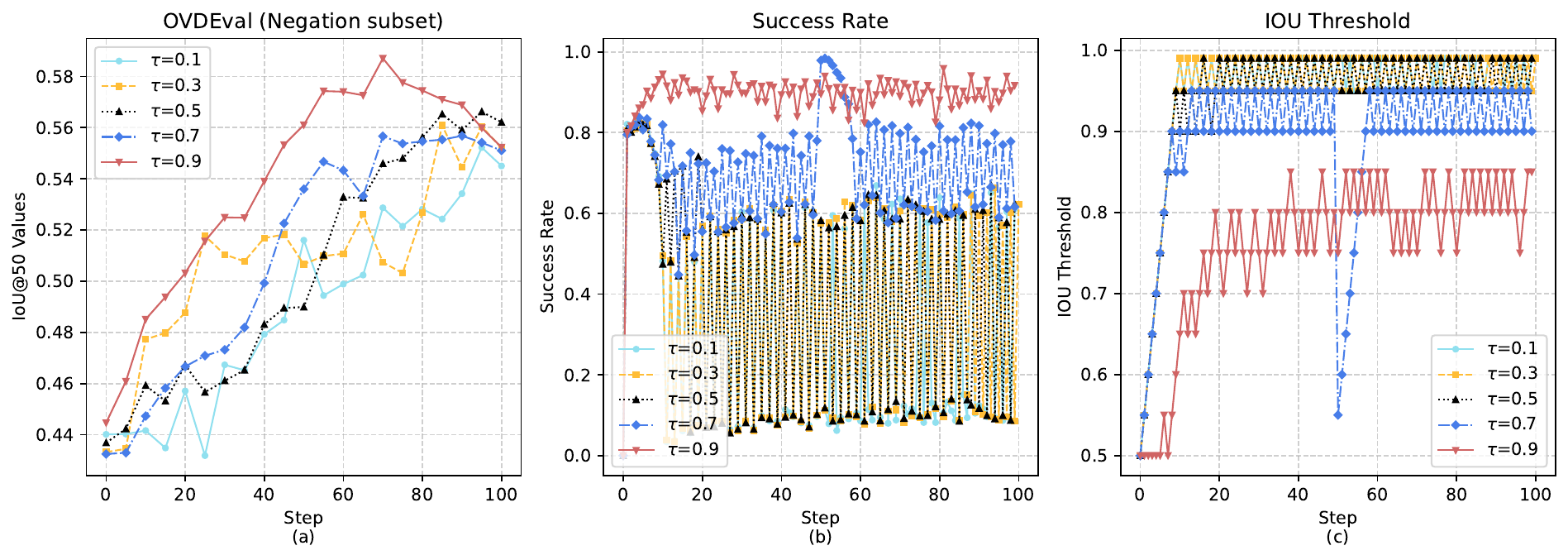}
    \caption{Training dynamics of adaptive IoU scheduling. We vary the target success rate $\tau \in \{0.1, \dots, 0.9\}$. (a) Validation performance (IoU@50) on OVDEval. (b) Batch Success Rate stability. (c) Evolution of the IoU threshold. The results highlight the trade-off between threshold stagnation at high $\tau$ and premature saturation at low $\tau$.}
    \label{fig:adaptive}
\end{figure}

We sweep $\tau \in \{0.1, 0.3, 0.5, 0.7, 0.9\}$ to cover controllers that are respectively more aggressive or more conservative in tightening the threshold.
The resulting training dynamics are shown in~\cref{fig:adaptive}.

\paragraph{Results.}
At high target success ($\tau=0.9$), the threshold remains relatively stable but tends to plateau around $0.85$, never reaching the strict high-precision regime.
At lower target success ($\tau \le 0.7$), the threshold rises to $0.99$ too early, which introduces severe reward sparsity and unstable training dynamics.
In other words, the adaptive controller does not remove the schedule-design problem, but shifts it to the choice of $\tau$.

\paragraph{Conclusion.}
While the adaptive scheduler is flexible, it introduces an additional control parameter that is not straightforward to calibrate.
In our setting, the fixed three-stage schedule remains the more stable and interpretable choice for driving training toward a high-precision regime.

%% file: sections/14_appx_mega_bench_curves.tex
\clearpage
\newpage
\section{MEGA-Bench Evaluation Curves}
\label{appx:mega_bench_curves}

We provide the full MEGA-Bench evaluation curves for the intermediate checkpoints of the 7B, 32B-0321, and 32B-0326 training runs.
The fourth panel further shows the task-level trajectory of Orsta-32B-0321 on MEGA-Bench.

\begin{figure}[ht]
    \centering
    \begin{subfigure}{0.24\textwidth}
        \includegraphics[width=\linewidth]{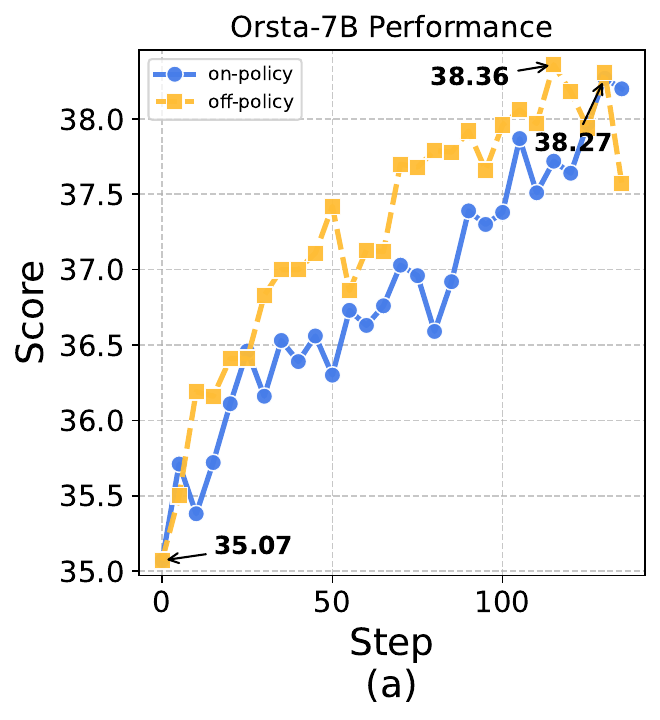}
    \end{subfigure}
    \begin{subfigure}{0.24\textwidth}
        \includegraphics[width=\linewidth]{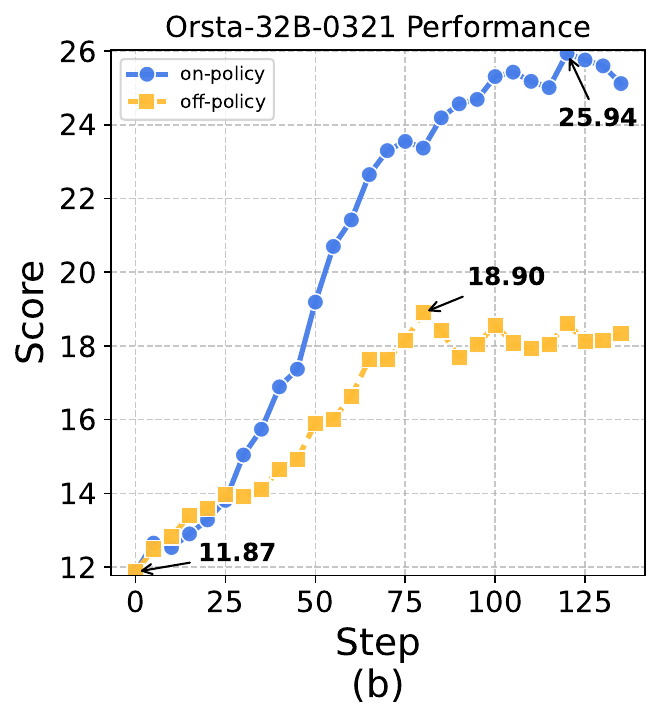}
    \end{subfigure}
    \begin{subfigure}{0.24\textwidth}
        \includegraphics[width=\linewidth]{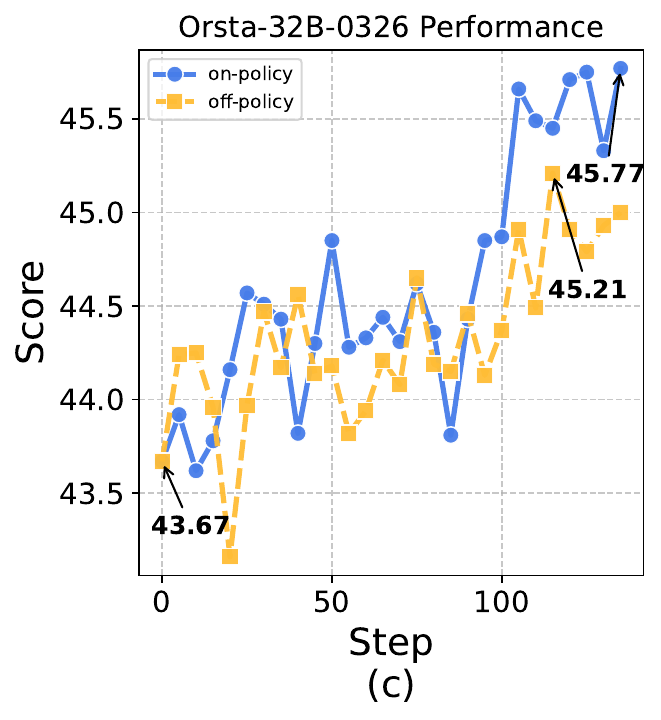}
    \end{subfigure}
    \begin{subfigure}{0.24\textwidth}
        \includegraphics[width=\linewidth]{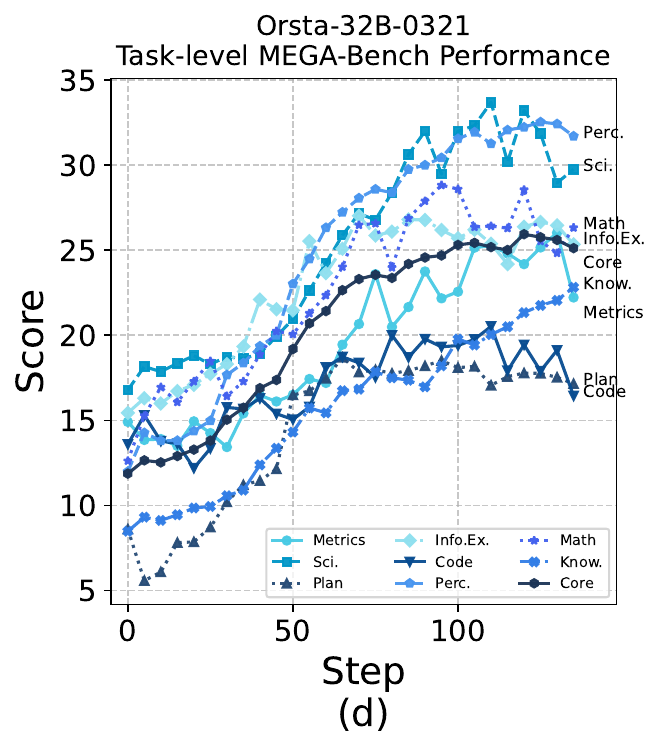}
    \end{subfigure}
    \caption{MEGA-Bench evaluation curves across training. The first three panels compare on-policy and off-policy variants for the 7B, 32B-0321, and 32B-0326 backbones. Models are evaluated every 5 generation steps from 0 to 135. The fourth panel shows the task-level trajectory of Orsta-32B-0321 on MEGA-Bench.}
    \label{fig:mega_performance_track}
\end{figure}

%% file: sections/15_appx_dynamic_iou_cases.tex
\clearpage
\newpage
\section{Qualitative Cases for Dynamic IoU}
\label{appx:dynamic_iou_cases}

We provide qualitative cases to illustrate the late-stage drift discussed in~\cref{subsec:dynamic_iou_ablation}.

\begin{figure*}[ht]
    \centering
    \includegraphics[width=0.8\linewidth]{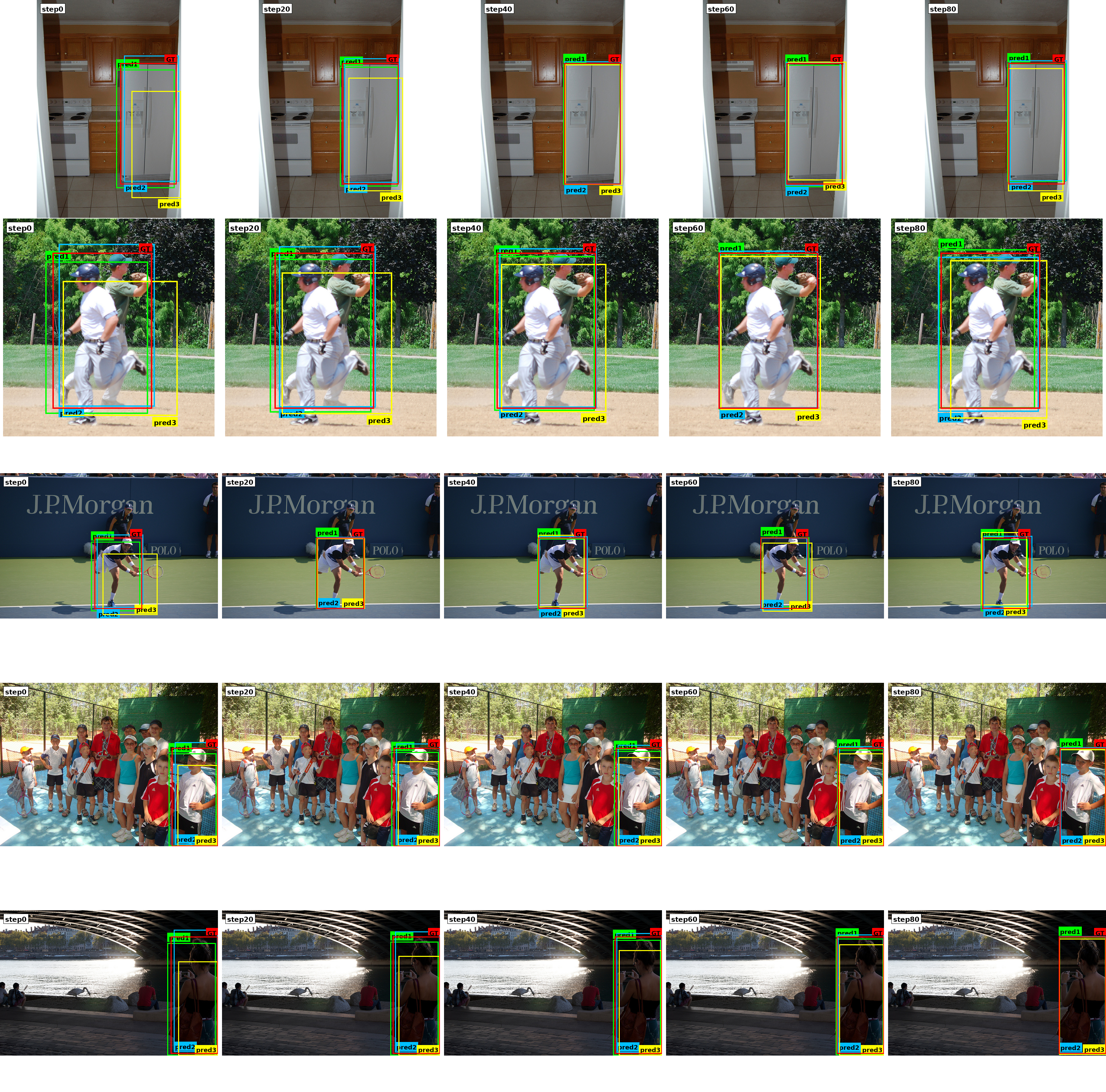}
    \caption{Qualitative cases from the fixed low-threshold ablation curve (IoU@50) shown in~\cref{fig:combined_iou_source_level}(a). Each row shows one COCO multi-object sample, where we select the largest ground-truth box in the image for visualization. Each column corresponds to an intermediate checkpoint along the IoU@50 training trajectory. In these cases, predictions become more accurate in the middle stage but later drift among multiple coarse boxes around the target, rather than continuing to sharpen around the ground-truth box. This pattern is consistent with the reward ambiguity induced by a loose threshold such as IoU@50, under which multiple coarse boxes can receive similarly high rewards.}
    \label{fig:dynamic_iou_cases}
\end{figure*}

These cases help explain the late-stage degradation of the IoU@50 curve in~\cref{fig:combined_iou_source_level}(a).
Although predictions become more accurate in the middle stage, later checkpoints often drift among several coarse boxes around the target instead of continuing to sharpen around the ground-truth box.
A loose threshold such as IoU@50 still provides non-zero reward through the IoU value itself, but it also creates a broad region in which multiple coarse predictions can receive similarly high rewards.
Once training enters this regime, the marginal reward difference between these boxes becomes weak, so optimization no longer strongly favors continued refinement toward the ground-truth box.
This qualitative pattern is consistent with the reward-ambiguity explanation used in the main text.

%% file: sections/16_appx_ovd_dynamic_iou.tex
\clearpage
\newpage
\section{OVDEval Curves for Dynamic IoU}
\label{appx:ovd_dynamic_iou}

We provide the OVDEval negation-subset curves referenced in~\cref{subsec:dynamic_iou_ablation}.
The comparison follows the same setup as the COCO ablations in the main text and includes fixed IoU@99 together with the three staged Dynamic IoU schedules.

\begin{figure*}[ht]
    \centering
    \includegraphics[width=0.7\linewidth]{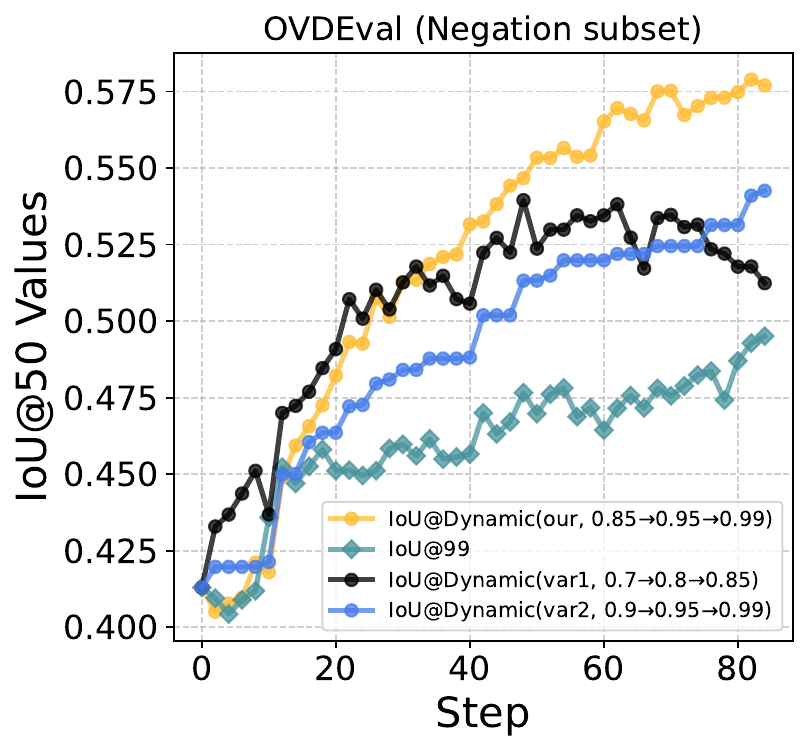}
    \caption{Dynamic IoU ablations on the OVDEval negation subset~\citep{yao2024evaluate}. The main schedule ($0.85 \rightarrow 0.95 \rightarrow 0.99$) outperforms fixed IoU@99 as well as the looser and stricter staged variants, matching the same overall pattern observed on COCO multi-object.}
    \label{fig:ovd_dynamic_iou}
\end{figure*}

%% file: main_v2.bbl
\begin{thebibliography}{57}
\providecommand{\natexlab}[1]{#1}
\providecommand{\url}[1]{\texttt{#1}}
\expandafter\ifx\csname urlstyle\endcsname\relax
  \providecommand{\doi}[1]{doi: #1}\else
  \providecommand{\doi}{doi: \begingroup \urlstyle{rm}\Url}\fi

\bibitem[{Apache Software Foundation}(2025)]{parquet}
{Apache Software Foundation}.
\newblock Apache parquet documentation.
\newblock \url{https://parquet.apache.org/docs/}, 2025.
\newblock Accessed: 2025-05-20.

\bibitem[Bai et~al.(2025)Bai, Chen, Liu, Wang, Ge, Song, Dang, Wang, Wang, Tang, Zhong, Zhu, Yang, Li, Wan, Wang, Ding, Fu, Xu, Ye, Zhang, Xie, Cheng, Zhang, Yang, Xu, and Lin]{Qwen2.5-VL}
Shuai Bai, Keqin Chen, Xuejing Liu, Jialin Wang, Wenbin Ge, Sibo Song, Kai Dang, Peng Wang, Shijie Wang, Jun Tang, Humen Zhong, Yuanzhi Zhu, Mingkun Yang, Zhaohai Li, Jianqiang Wan, Pengfei Wang, Wei Ding, Zheren Fu, Yiheng Xu, Jiabo Ye, Xi~Zhang, Tianbao Xie, Zesen Cheng, Hang Zhang, Zhibo Yang, Haiyang Xu, and Junyang Lin.
\newblock Qwen2.5-vl technical report.
\newblock \emph{arXiv preprint arXiv:2502.13923}, 2025.

\bibitem[Borisova and Rehm(2025)]{borisova2025scivqa}
Ekaterina Borisova and Georg Rehm.
\newblock Scivqa: Scientific visual question answering.
\newblock \emph{SDProc 2025}, 2025.
\newblock URL \url{https://sdproc.org/2025/scivqa.html}.

\bibitem[Chen et~al.(2024)Chen, Liang, Siu, Wang, Wang, Wang, Ni, Zhu, Jiang, Lyu, et~al.]{chen2024mega}
Jiacheng Chen, Tianhao Liang, Sherman Siu, Zhengqing Wang, Kai Wang, Yubo Wang, Yuansheng Ni, Wang Zhu, Ziyan Jiang, Bohan Lyu, et~al.
\newblock Mega-bench: Scaling multimodal evaluation to over 500 real-world tasks.
\newblock \emph{arXiv preprint arXiv:2410.10563}, 2024.

\bibitem[Chia et~al.(2024)Chia, Han, Ghosal, Bing, and Poria]{chia2024puzzlevqa}
Yew~Ken Chia, Vernon Toh~Yan Han, Deepanway Ghosal, Lidong Bing, and Soujanya Poria.
\newblock Puzzlevqa: Diagnosing multimodal reasoning challenges of language models with abstract visual patterns.
\newblock \emph{arXiv preprint arXiv:2403.13315}, 2024.

\bibitem[Duan et~al.(2024)Duan, Yang, Qiao, Fang, Chen, Liu, Dong, Zang, Zhang, Wang, et~al.]{duan2024vlmevalkit}
Haodong Duan, Junming Yang, Yuxuan Qiao, Xinyu Fang, Lin Chen, Yuan Liu, Xiaoyi Dong, Yuhang Zang, Pan Zhang, Jiaqi Wang, et~al.
\newblock Vlmevalkit: An open-source toolkit for evaluating large multi-modality models.
\newblock In \emph{Proceedings of the 32nd ACM International Conference on Multimedia}, pages 11198--11201, 2024.

\bibitem[Feng et~al.(2025)Feng, Zhang, Li, Fan, Chen, Jiang, Zheng, Sun, Zhang, Sun, et~al.]{feng2025onethinker}
Kaituo Feng, Manyuan Zhang, Hongyu Li, Kaixuan Fan, Shuang Chen, Yilei Jiang, Dian Zheng, Peiwen Sun, Yiyuan Zhang, Haoze Sun, et~al.
\newblock Onethinker: All-in-one reasoning model for image and video.
\newblock \emph{arXiv preprint arXiv:2512.03043}, 2025.

\bibitem[Fu et~al.(2024)Fu, Yang, Kuang, Song, Li, Zhu, Luo, Wang, Lu, Huang, et~al.]{fu2024ocrbench}
Ling Fu, Biao Yang, Zhebin Kuang, Jiajun Song, Yuzhe Li, Linghao Zhu, Qidi Luo, Xinyu Wang, Hao Lu, Mingxin Huang, et~al.
\newblock Ocrbench v2: An improved benchmark for evaluating large multimodal models on visual text localization and reasoning.
\newblock \emph{arXiv preprint arXiv:2501.00321}, 2024.

\bibitem[Ghosal et~al.(2024)Ghosal, Han, Chia, and Poria]{ghosal2024algopuzzlevqa}
Deepanway Ghosal, Vernon Toh~Yan Han, Yew~Ken Chia, and Soujanya Poria.
\newblock Are language models puzzle prodigies? algorithmic puzzles unveil serious challenges in multimodal reasoning.
\newblock \emph{arXiv preprint arXiv:2403.03864}, 2024.

\bibitem[Huang et~al.(2025)Huang, Jia, Zhai, Cao, Ye, Zhao, Xu, Hu, and Lin]{huang2025vision}
Wenxuan Huang, Bohan Jia, Zijie Zhai, Shaosheng Cao, Zheyu Ye, Fei Zhao, Zhe Xu, Yao Hu, and Shaohui Lin.
\newblock Vision-r1: Incentivizing reasoning capability in multimodal large language models.
\newblock \emph{arXiv preprint arXiv:2503.06749}, 2025.

\bibitem[Johnson et~al.(2017)Johnson, Hariharan, Van Der~Maaten, Fei-Fei, Lawrence~Zitnick, and Girshick]{johnson2017clevr}
Justin Johnson, Bharath Hariharan, Laurens Van Der~Maaten, Li~Fei-Fei, C~Lawrence~Zitnick, and Ross Girshick.
\newblock Clevr: A diagnostic dataset for compositional language and elementary visual reasoning.
\newblock In \emph{Proceedings of the IEEE conference on computer vision and pattern recognition}, pages 2901--2910, 2017.

\bibitem[Kim et~al.(2024)Kim, Yim, and Song]{kim2024tablevqa}
Yoonsik Kim, Moonbin Yim, and Ka~Yeon Song.
\newblock Tablevqa-bench: A visual question answering benchmark on multiple table domains.
\newblock \emph{arXiv preprint arXiv:2404.19205}, 2024.

\bibitem[Kimi et~al.(2025)Kimi, Du, Yin, Xing, Qu, Wang, Chen, Zhang, Du, Wei, Wang, Zhang, Du, Wang, Yuan, Lu, Li, Sung, Wei, Lai, Zhu, Ding, Hu, Yang, Zhang, Wu, Yao, Lu, Wang, Gao, Zheng, Li, Su, Wang, Deng, Qiu, Xie, Wang, Liu, Yan, Ouyang, Chen, Sui, Yu, Dong, Dong, Xu, Cheng, Gu, Zhou, Liu, Cao, Yu, Song, Bai, Song, He, Huang, Xu, Yuan, Yao, Wu, Zu, Zhou, Wang, Charles, Zhong, Li, Hu, Chen, Wang, Liu, Miao, Qin, Chen, Bao, Wang, Kang, Liu, Du, Wu, Wang, Yan, Zhou, Li, Jiang, Zhang, Yang, Huang, Huang, Zhao, and Chen]{kimiteam2025kimivltechnicalreport}
Team Kimi, Angang Du, Bohong Yin, Bowei Xing, Bowen Qu, Bowen Wang, Cheng Chen, Chenlin Zhang, Chenzhuang Du, Chu Wei, Congcong Wang, Dehao Zhang, Dikang Du, Dongliang Wang, Enming Yuan, Enzhe Lu, Fang Li, Flood Sung, Guangda Wei, Guokun Lai, Han Zhu, Hao Ding, Hao Hu, Hao Yang, Hao Zhang, Haoning Wu, Haotian Yao, Haoyu Lu, Heng Wang, Hongcheng Gao, Huabin Zheng, Jiaming Li, Jianlin Su, Jianzhou Wang, Jiaqi Deng, Jiezhong Qiu, Jin Xie, Jinhong Wang, Jingyuan Liu, Junjie Yan, Kun Ouyang, Liang Chen, Lin Sui, Longhui Yu, Mengfan Dong, Mengnan Dong, Nuo Xu, Pengyu Cheng, Qizheng Gu, Runjie Zhou, Shaowei Liu, Sihan Cao, Tao Yu, Tianhui Song, Tongtong Bai, Wei Song, Weiran He, Weixiao Huang, Weixin Xu, Xiaokun Yuan, Xingcheng Yao, Xingzhe Wu, Xinxing Zu, Xinyu Zhou, Xinyuan Wang, Y.~Charles, Yan Zhong, Yang Li, Yangyang Hu, Yanru Chen, Yejie Wang, Yibo Liu, Yibo Miao, Yidao Qin, Yimin Chen, Yiping Bao, Yiqin Wang, Yongsheng Kang, Yuanxin Liu, Yulun Du, Yuxin Wu, Yuzhi Wang, Yuzi Yan, Zaida Zhou, Zhaowei Li, Zhejun
  Jiang, Zheng Zhang, Zhilin Yang, Zhiqi Huang, Zihao Huang, Zijia Zhao, and Ziwei Chen.
\newblock {Kimi-VL} technical report, 2025.
\newblock URL \url{https://arxiv.org/abs/2504.07491}.

\bibitem[Kimi et~al.(2026)Kimi, Bai, Bai, Bao, Cai, Cao, Charles, Che, Chen, Chen, et~al.]{team2026kimi}
Team Kimi, Tongtong Bai, Yifan Bai, Yiping Bao, SH~Cai, Yuan Cao, Y~Charles, HS~Che, Cheng Chen, Guanduo Chen, et~al.
\newblock Kimi k2. 5: Visual agentic intelligence.
\newblock \emph{arXiv preprint arXiv:2602.02276}, 2026.

\bibitem[Kydlíček(2025)]{mathverify2025}
Hynek Kydlíček.
\newblock Math-verify: A library for rule-based verification of mathematical answers, 2025.
\newblock URL \url{https://github.com/huggingface/Math-Verify}.

\bibitem[Li et~al.(2024)Li, Zhang, Guo, Zhang, Li, Zhang, Zhang, Zhang, Li, Liu, et~al.]{li2024llava}
Bo~Li, Yuanhan Zhang, Dong Guo, Renrui Zhang, Feng Li, Hao Zhang, Kaichen Zhang, Peiyuan Zhang, Yanwei Li, Ziwei Liu, et~al.
\newblock Llava-onevision: Easy visual task transfer.
\newblock \emph{arXiv preprint arXiv:2408.03326}, 2024.

\bibitem[Li et~al.(2025)Li, Meng, Lin, Luo, Tian, Ma, Huang, and Chua]{li2025screenspot}
Kaixin Li, Ziyang Meng, Hongzhan Lin, Ziyang Luo, Yuchen Tian, Jing Ma, Zhiyong Huang, and Tat-Seng Chua.
\newblock Screenspot-pro: Gui grounding for professional high-resolution computer use.
\newblock \emph{arXiv preprint arXiv:2504.07981}, 2025.

\bibitem[Lin et~al.(2025)Lin, Li, Gao, Yang, Wu, Bai, Lei, Wang, and Shou]{lin2025showui}
Kevin~Qinghong Lin, Linjie Li, Difei Gao, Zhengyuan Yang, Shiwei Wu, Zechen Bai, Stan~Weixian Lei, Lijuan Wang, and Mike~Zheng Shou.
\newblock Showui: One vision-language-action model for gui visual agent.
\newblock In \emph{Proceedings of the Computer Vision and Pattern Recognition Conference}, pages 19498--19508, 2025.

\bibitem[Lin et~al.(2014)Lin, Maire, Belongie, Hays, Perona, Ramanan, Doll{\'a}r, and Zitnick]{lin2014microsoft}
Tsung-Yi Lin, Michael Maire, Serge Belongie, James Hays, Pietro Perona, Deva Ramanan, Piotr Doll{\'a}r, and C~Lawrence Zitnick.
\newblock Microsoft coco: Common objects in context.
\newblock In \emph{Computer vision--ECCV 2014: 13th European conference, zurich, Switzerland, September 6-12, 2014, proceedings, part v 13}, pages 740--755. Springer, 2014.

\bibitem[Liu et~al.(2025{\natexlab{a}})Liu, Mei, Lin, Xue, Wang, Xu, Wu, Zhang, Lin, Dong, et~al.]{liu2025deepseek}
Aixin Liu, Aoxue Mei, Bangcai Lin, Bing Xue, Bingxuan Wang, Bingzheng Xu, Bochao Wu, Bowei Zhang, Chaofan Lin, Chen Dong, et~al.
\newblock Deepseek-v3. 2: Pushing the frontier of open large language models.
\newblock \emph{arXiv preprint arXiv:2512.02556}, 2025{\natexlab{a}}.

\bibitem[Liu et~al.(2025{\natexlab{b}})Liu, Qu, Zhong, Peng, Liu, Yu, and Jia]{liu2025visionreasoner}
Yuqi Liu, Tianyuan Qu, Zhisheng Zhong, Bohao Peng, Shu Liu, Bei Yu, and Jiaya Jia.
\newblock Visionreasoner: Unified visual perception and reasoning via reinforcement learning.
\newblock \emph{arXiv preprint arXiv:2505.12081}, 2025{\natexlab{b}}.

\bibitem[Liu et~al.(2025{\natexlab{c}})Liu, Zhang, Liu, Zhang, Sun, and Wang]{liu2025othink}
Zhiyuan Liu, Yuting Zhang, Feng Liu, Changwang Zhang, Ying Sun, and Jun Wang.
\newblock Othink-mr1: Stimulating multimodal generalized reasoning capabilities via dynamic reinforcement learning.
\newblock \emph{arXiv preprint arXiv:2503.16081}, 2025{\natexlab{c}}.

\bibitem[Liu et~al.(2025{\natexlab{d}})Liu, Sun, Zang, Dong, Cao, Duan, Lin, and Wang]{liu2025visual}
Ziyu Liu, Zeyi Sun, Yuhang Zang, Xiaoyi Dong, Yuhang Cao, Haodong Duan, Dahua Lin, and Jiaqi Wang.
\newblock Visual-rft: Visual reinforcement fine-tuning.
\newblock \emph{arXiv preprint arXiv:2503.01785}, 2025{\natexlab{d}}.

\bibitem[Lu et~al.(2021)Lu, Gong, Jiang, Qiu, Huang, Liang, and Zhu]{lu2021inter}
Pan Lu, Ran Gong, Shibiao Jiang, Liang Qiu, Siyuan Huang, Xiaodan Liang, and Song-Chun Zhu.
\newblock Inter-gps: Interpretable geometry problem solving with formal language and symbolic reasoning.
\newblock \emph{arXiv preprint arXiv:2105.04165}, 2021.

\bibitem[Lu et~al.(2022)Lu, Mishra, Xia, Qiu, Chang, Zhu, Tafjord, Clark, and Kalyan]{lu2022learn}
Pan Lu, Swaroop Mishra, Tony Xia, Liang Qiu, Kai-Wei Chang, Song-Chun Zhu, Oyvind Tafjord, Peter Clark, and Ashwin Kalyan.
\newblock Learn to explain: Multimodal reasoning via thought chains for science question answering.
\newblock In \emph{The 36th Conference on Neural Information Processing Systems (NeurIPS)}, 2022.

\bibitem[Lu et~al.(2023)Lu, Bansal, Xia, Liu, Li, Hajishirzi, Cheng, Chang, Galley, and Gao]{lu2023mathvista}
Pan Lu, Hritik Bansal, Tony Xia, Jiacheng Liu, Chunyuan Li, Hannaneh Hajishirzi, Hao Cheng, Kai-Wei Chang, Michel Galley, and Jianfeng Gao.
\newblock Mathvista: Evaluating math reasoning in visual contexts with gpt-4v, bard, and other large multimodal models.
\newblock \emph{CoRR}, 2023.

\bibitem[Ma et~al.(2025{\natexlab{a}})Ma, Ding, Luo, Chen, Guo, Wong, Feng, and Sun]{ma2025deepperception}
Xinyu Ma, Ziyang Ding, Zhicong Luo, Chi Chen, Zonghao Guo, Derek~F Wong, Xiaoyi Feng, and Maosong Sun.
\newblock Deepperception: Advancing r1-like cognitive visual perception in mllms for knowledge-intensive visual grounding.
\newblock \emph{arXiv preprint arXiv:2503.12797}, 2025{\natexlab{a}}.

\bibitem[Ma et~al.(2025{\natexlab{b}})Ma, Chern, Shen, Zhong, and Liu]{ma2025rethinking}
Yan Ma, Steffi Chern, Xuyang Shen, Yiran Zhong, and Pengfei Liu.
\newblock Rethinking rl scaling for vision language models: A transparent, from-scratch framework and comprehensive evaluation scheme.
\newblock \emph{arXiv preprint arXiv:2504.02587}, 2025{\natexlab{b}}.

\bibitem[Masry et~al.(2025)Masry, Islam, Ahmed, Bajaj, Kabir, Kartha, Laskar, Rahman, Rahman, Shahmohammadi, et~al.]{masry2025chartqapro}
Ahmed Masry, Mohammed~Saidul Islam, Mahir Ahmed, Aayush Bajaj, Firoz Kabir, Aaryaman Kartha, Md~Tahmid~Rahman Laskar, Mizanur Rahman, Shadikur Rahman, Mehrad Shahmohammadi, et~al.
\newblock Chartqapro: A more diverse and challenging benchmark for chart question answering.
\newblock \emph{arXiv preprint arXiv:2504.05506}, 2025.

\bibitem[Meituan~LongCat et~al.(2025)Meituan~LongCat, Li, Lei, Wang, Rong, Wang, Zhang, Gao, Zhang, Sun, et~al.]{team2025longcat}
Team Meituan~LongCat, Bei Li, Bingye Lei, Bo~Wang, Bolin Rong, Chao Wang, Chao Zhang, Chen Gao, Chen Zhang, Cheng Sun, et~al.
\newblock Longcat-flash technical report.
\newblock \emph{arXiv preprint arXiv:2509.01322}, 2025.

\bibitem[Meng et~al.(2025)Meng, Du, Liu, Zhou, Lu, Fu, Han, Shi, Wang, He, et~al.]{meng2025mm}
Fanqing Meng, Lingxiao Du, Zongkai Liu, Zhixiang Zhou, Quanfeng Lu, Daocheng Fu, Tiancheng Han, Botian Shi, Wenhai Wang, Junjun He, et~al.
\newblock Mm-eureka: Exploring the frontiers of multimodal reasoning with rule-based reinforcement learning.
\newblock \emph{arXiv preprint arXiv:2503.07365}, 2025.

\bibitem[{Qwen Team}(2026)]{qwen3.5}
{Qwen Team}.
\newblock {Qwen3.5}: Towards native multimodal agents, February 2026.
\newblock URL \url{https://qwen.ai/blog?id=qwen3.5}.

\bibitem[Shao et~al.(2019)Shao, Li, Zhang, Peng, Yu, Zhang, Li, and Sun]{shao2019objects365}
Shuai Shao, Zeming Li, Tianyuan Zhang, Chao Peng, Gang Yu, Xiangyu Zhang, Jing Li, and Jian Sun.
\newblock Objects365: A large-scale, high-quality dataset for object detection.
\newblock In \emph{Proceedings of the IEEE/CVF international conference on computer vision}, pages 8430--8439, 2019.

\bibitem[Shao et~al.(2024)Shao, Wang, Zhu, Xu, Song, Bi, Zhang, Zhang, Li, Wu, et~al.]{shao2024deepseekmath}
Zhihong Shao, Peiyi Wang, Qihao Zhu, Runxin Xu, Junxiao Song, Xiao Bi, Haowei Zhang, Mingchuan Zhang, YK~Li, Y~Wu, et~al.
\newblock Deepseekmath: Pushing the limits of mathematical reasoning in open language models.
\newblock \emph{arXiv preprint arXiv:2402.03300}, 2024.

\bibitem[Shen et~al.(2025)Shen, Liu, Li, Fang, Ma, Liao, Shen, Zhang, Zhao, Zhang, et~al.]{shen2025vlm}
Haozhan Shen, Peng Liu, Jingcheng Li, Chunxin Fang, Yibo Ma, Jiajia Liao, Qiaoli Shen, Zilun Zhang, Kangjia Zhao, Qianqian Zhang, et~al.
\newblock Vlm-r1: A stable and generalizable r1-style large vision-language model.
\newblock \emph{arXiv preprint arXiv:2504.07615}, 2025.

\bibitem[Sheng et~al.(2024)Sheng, Zhang, Ye, Wu, Zhang, Zhang, Peng, Lin, and Wu]{sheng2024hybridflow}
Guangming Sheng, Chi Zhang, Zilingfeng Ye, Xibin Wu, Wang Zhang, Ru~Zhang, Yanghua Peng, Haibin Lin, and Chuan Wu.
\newblock Hybridflow: A flexible and efficient rlhf framework.
\newblock \emph{arXiv preprint arXiv: 2409.19256}, 2024.

\bibitem[Song et~al.(2025)Song, Ou, Kong, Li, Neubig, and Yue]{song2025visualpuzzles}
Yueqi Song, Tianyue Ou, Yibo Kong, Zecheng Li, Graham Neubig, and Xiang Yue.
\newblock Visualpuzzles: Decoupling multimodal reasoning evaluation from domain knowledge.
\newblock \emph{arXiv preprint arXiv:2504.10342}, 2025.

\bibitem[Sun et~al.(2024)Sun, Bai, Qi, Hou, and Li]{sun2024mm}
Kai Sun, Yushi Bai, Ji~Qi, Lei Hou, and Juanzi Li.
\newblock Mm-math: Advancing multimodal math evaluation with process evaluation and fine-grained classification.
\newblock \emph{arXiv preprint arXiv:2404.05091}, 2024.

\bibitem[Tan et~al.(2025)Tan, Ji, Hao, Lin, Wang, Wang, and Zhang]{tan2025reason}
Huajie Tan, Yuheng Ji, Xiaoshuai Hao, Minglan Lin, Pengwei Wang, Zhongyuan Wang, and Shanghang Zhang.
\newblock Reason-rft: Reinforcement fine-tuning for visual reasoning.
\newblock \emph{arXiv preprint arXiv:2503.20752}, 2025.

\bibitem[Wang et~al.(2025{\natexlab{a}})Wang, Qu, Huang, Chu, Lin, and Chen]{vl_rethinker}
Haozhe Wang, Chao Qu, Zuming Huang, Wei Chu, Fangzhen Lin, and Wenhu Chen.
\newblock Vl-rethinker: Incentivizing self-reflection of vision-language models with reinforcement learning.
\newblock \emph{arXiv preprint arXiv:2504.08837}, 2025{\natexlab{a}}.

\bibitem[Wang et~al.(2025{\natexlab{b}})Wang, Qu, Huang, Chu, Lin, and Chen]{wang2025vl}
Haozhe Wang, Chao Qu, Zuming Huang, Wei Chu, Fangzhen Lin, and Wenhu Chen.
\newblock Vl-rethinker: Incentivizing self-reflection of vision-language models with reinforcement learning.
\newblock \emph{arXiv preprint arXiv:2504.08837}, 2025{\natexlab{b}}.

\bibitem[Wang et~al.(2024{\natexlab{a}})Wang, Pan, Shi, Lu, Ren, Zhou, Zhan, and Li]{mathvision}
Ke~Wang, Junting Pan, Weikang Shi, Zimu Lu, Houxing Ren, Aojun Zhou, Mingjie Zhan, and Hongsheng Li.
\newblock Measuring multimodal mathematical reasoning with math-vision dataset.
\newblock In \emph{The Thirty-eight Conference on Neural Information Processing Systems Datasets and Benchmarks Track}, 2024{\natexlab{a}}.
\newblock URL \url{https://openreview.net/forum?id=QWTCcxMpPA}.

\bibitem[Wang et~al.(2024{\natexlab{b}})Wang, Ding, Zeng, Zhou, Shen, Luo, and Tao]{hrbench}
Wenbin Wang, Liang Ding, Minyan Zeng, Xiabin Zhou, Li~Shen, Yong Luo, and Dacheng Tao.
\newblock Divide, conquer and combine: A training-free framework for high-resolution image perception in multimodal large language models.
\newblock \emph{arXiv preprint}, 2024{\natexlab{b}}.
\newblock URL \url{https://arxiv.org/abs/2408.15556}.

\bibitem[Wang et~al.(2020)Wang, Liu, Shen, Ng, Luo, Jin, Chan, Hengel, and Wang]{wang2020general}
Xinyu Wang, Yuliang Liu, Chunhua Shen, Chun~Chet Ng, Canjie Luo, Lianwen Jin, Chee~Seng Chan, Anton van~den Hengel, and Liangwei Wang.
\newblock On the general value of evidence, and bilingual scene-text visual question answering.
\newblock In \emph{Proceedings of the IEEE/CVF Conference on Computer Vision and Pattern Recognition}, pages 10126--10135, 2020.

\bibitem[Wang et~al.(2024{\natexlab{c}})Wang, Xia, He, Chen, Liu, Zhu, Liang, Wu, Liu, Malladi, et~al.]{wang2024charxiv}
Zirui Wang, Mengzhou Xia, Luxi He, Howard Chen, Yitao Liu, Richard Zhu, Kaiqu Liang, Xindi Wu, Haotian Liu, Sadhika Malladi, et~al.
\newblock Charxiv: Charting gaps in realistic chart understanding in multimodal llms.
\newblock \emph{Advances in Neural Information Processing Systems}, 37:\penalty0 113569--113697, 2024{\natexlab{c}}.

\bibitem[Wu and Xie(2023)]{vstar}
Penghao Wu and Saining Xie.
\newblock V*: Guided visual search as a core mechanism in multimodal llms.
\newblock \emph{arXiv preprint arXiv:2312.14135}, 2023.

\bibitem[Xia et~al.(2024)Xia, Zhang, Ye, Yan, Liu, Zhou, Chen, Ye, Dou, Shi, et~al.]{xia2024chartx}
Renqiu Xia, Bo~Zhang, Hancheng Ye, Xiangchao Yan, Qi~Liu, Hongbin Zhou, Zijun Chen, Peng Ye, Min Dou, Botian Shi, et~al.
\newblock Chartx \& chartvlm: A versatile benchmark and foundation model for complicated chart reasoning.
\newblock \emph{arXiv preprint arXiv:2402.12185}, 2024.

\bibitem[Xiao et~al.(2026)Xiao, Xia, Yang, Gao, Shen, Zhang, He, Lou, Luo, Wang, et~al.]{xiao2026mimo}
Bangjun Xiao, Bingquan Xia, Bo~Yang, Bofei Gao, Bowen Shen, Chen Zhang, Chenhong He, Chiheng Lou, Fuli Luo, Gang Wang, et~al.
\newblock Mimo-v2-flash technical report.
\newblock \emph{arXiv preprint arXiv:2601.02780}, 2026.

\bibitem[Xie et~al.(2023)Xie, Zhang, Wu, Zhu, Zhao, and Liang]{xie2023described}
Chi Xie, Zhao Zhang, Yixuan Wu, Feng Zhu, Rui Zhao, and Shuang Liang.
\newblock Described object detection: Liberating object detection with flexible expressions.
\newblock \emph{Advances in Neural Information Processing Systems}, 36:\penalty0 79095--79107, 2023.

\bibitem[Xu et~al.(2025)Xu, Li, Yang, Zhang, Sun, Chow, Li, Song, Xu, Tong, et~al.]{xu2025mixedr1}
Shilin Xu, Yanwei Li, Rui Yang, Tao Zhang, Yueyi Sun, Wei Chow, Linfeng Li, Hang Song, Qi~Xu, Yunhai Tong, et~al.
\newblock Mixed-r1: Unified reward perspective for reasoning capability in multimodal large language models.
\newblock \emph{arXiv preprint arXiv:2505.24164}, 2025.

\bibitem[Yang et~al.(2025)Yang, He, Pan, Jiang, Deng, Yang, Lu, Yin, Rao, Zhu, et~al.]{yang2025r1}
Yi~Yang, Xiaoxuan He, Hongkun Pan, Xiyan Jiang, Yan Deng, Xingtao Yang, Haoyu Lu, Dacheng Yin, Fengyun Rao, Minfeng Zhu, et~al.
\newblock R1-onevision: Advancing generalized multimodal reasoning through cross-modal formalization.
\newblock \emph{arXiv preprint arXiv:2503.10615}, 2025.

\bibitem[Yao et~al.(2024)Yao, Liu, Zhao, Zhang, Liao, Fang, Lee, and Wang]{yao2024evaluate}
Yiyang Yao, Peng Liu, Tiancheng Zhao, Qianqian Zhang, Jiajia Liao, Chunxin Fang, Kyusong Lee, and Qing Wang.
\newblock How to evaluate the generalization of detection? a benchmark for comprehensive open-vocabulary detection.
\newblock In \emph{Proceedings of the AAAI Conference on Artificial Intelligence}, volume~38, pages 6630--6638, 2024.

\bibitem[Yu et~al.(2025)Yu, Lin, Zhao, Yin, Wei, Peng, Wei, Sun, Han, Ge, et~al.]{yu2025perception}
En~Yu, Kangheng Lin, Liang Zhao, Jisheng Yin, Yana Wei, Yuang Peng, Haoran Wei, Jianjian Sun, Chunrui Han, Zheng Ge, et~al.
\newblock Perception-r1: Pioneering perception policy with reinforcement learning.
\newblock \emph{arXiv preprint arXiv:2504.07954}, 2025.

\bibitem[Yuan et~al.(2025)Yuan, Peng, Jiang, Lu, Zhang, Feng, Fu, Chen, Bai, Zhang, et~al.]{yuan2025mme}
Jiakang Yuan, Tianshuo Peng, Yilei Jiang, Yiting Lu, Renrui Zhang, Kaituo Feng, Chaoyou Fu, Tao Chen, Lei Bai, Bo~Zhang, et~al.
\newblock Mme-reasoning: A comprehensive benchmark for logical reasoning in mllms.
\newblock \emph{arXiv preprint arXiv:2505.21327}, 2025.

\bibitem[Yue et~al.(2024)Yue, Ni, Zhang, Zheng, Liu, Zhang, Stevens, Jiang, Ren, Sun, et~al.]{yue2024mmmu}
Xiang Yue, Yuansheng Ni, Kai Zhang, Tianyu Zheng, Ruoqi Liu, Ge~Zhang, Samuel Stevens, Dongfu Jiang, Weiming Ren, Yuxuan Sun, et~al.
\newblock Mmmu: A massive multi-discipline multimodal understanding and reasoning benchmark for expert agi.
\newblock In \emph{Proceedings of the IEEE/CVF Conference on Computer Vision and Pattern Recognition}, pages 9556--9567, 2024.

\bibitem[Zhan et~al.(2025)Zhan, Wu, Zhu, Xue, Luo, Chen, Zhang, Li, He, Yang, et~al.]{zhan2025gthinker}
Yufei Zhan, Ziheng Wu, Yousong Zhu, Rongkun Xue, Ruipu Luo, Zhenghao Chen, Can Zhang, Yifan Li, Zhentao He, Zheming Yang, et~al.
\newblock Gthinker: Towards general multimodal reasoning via cue-guided rethinking.
\newblock \emph{arXiv preprint arXiv:2506.01078}, 2025.

\bibitem[Zhang et~al.(2024)Zhang, Li, Zhang, Pu, Cahyono, Hu, Liu, Zhang, Yang, Li, and Liu]{zhang2024lmmsevalrealitycheckevaluation}
Kaichen Zhang, Bo~Li, Peiyuan Zhang, Fanyi Pu, Joshua~Adrian Cahyono, Kairui Hu, Shuai Liu, Yuanhan Zhang, Jingkang Yang, Chunyuan Li, and Ziwei Liu.
\newblock Lmms-eval: Reality check on the evaluation of large multimodal models, 2024.
\newblock URL \url{https://arxiv.org/abs/2407.12772}.

\end{thebibliography}
